\documentclass[journal]{IEEEtran}
\usepackage{cite}
\usepackage{amsmath,amssymb,amsfonts}
\usepackage{graphicx}
\usepackage{textcomp}
\usepackage{url}
\usepackage{multirow}
\begin{document}

\title{RadarGSOcc: Doppler-Aware 3D Semantic Occupancy Prediction Using 4D Radar Tensor with Geometric-Semantic Feature Extraction}

\author{Christofel Rio Goenawan, Dong-Hee Paek,~\IEEEmembership{Member,~IEEE}, and Seung-Hyun Kong,~\IEEEmembership{Senior~Member,~IEEE}
\thanks{This work has been submitted to the IEEE for possible publication. Copyright may be transferred without notice, after which this version may no longer be accessible.}
\thanks{Christofel Rio Goenawan is with the Robotics Program, Korea Advanced Institute of Science and Technology (KAIST), Daejeon, Korea (e-mail: christofel.goenawan@kaist.ac.kr).}
\thanks{Dong-Hee Paek was with the CCS Graduate School of Mobility, KAIST, Daejeon, Korea. He is now with the Department of Future Mobility, Korea University, Sejong, Korea (e-mail: dhpaek@korea.ac.kr).}
\thanks{Seung-Hyun Kong is with the CCS Graduate School of Mobility, KAIST, Daejeon, Korea (e-mail: skong@kaist.ac.kr). (\textit{Corresponding author: Seung-Hyun Kong.})}}

\markboth{Goenawan \MakeLowercase{\textit{et al.}}: RadarGSOcc: Doppler-Aware 3D Semantic Occupancy Prediction Using 4D Radar Tensor}%
{Goenawan \MakeLowercase{\textit{et al.}}: RadarGSOcc: Doppler-Aware 3D Semantic Occupancy Prediction Using 4D Radar Tensor}

\maketitle

\begin{abstract}
3D semantic occupancy prediction is crucial for autonomous driving, but camera-based methods degrade in low-light conditions and LiDAR-based methods are sensitive to adverse weather. 4D radar is more robust, yet its sparsity and the high cost of radar tensor processing limit its use. We propose Radar Geometric-Semantic Occupancy Prediction (RadarGSOcc), an efficient framework that directly processes 4D radar tensors by separately extracting geometric and semantic features, including Doppler information, with a deformable attention-based geometric refinement module as its key component. We also introduce the K-Radar 3DSemOcc benchmark, the first large-scale benchmark for fine-grained 3D semantic occupancy prediction using 4D radar tensors, built on the K-Radar dataset with refined LiDAR-based annotations. RadarGSOcc predicts 15 classes within a 51.2\,m range and achieves 38.2\% IoU, 7.6\% mIoU, and 25.5\% weighted mIoU, a 16\% relative improvement over RadarOcc in weighted mIoU (25.5 vs.\ 21.9; 23.8 vs.\ 20.9 averaged over three runs) with fewer parameters and FLOPs. The gain comes mainly from large static classes such as drivable area, vegetation, and terrain, whereas vehicles remain difficult: RadarOcc is better when car and truck/bus are merged into one class. RadarGSOcc also outperforms camera-based and camera--radar fusion baselines and narrows the gap with LiDAR-based methods. To assess robustness in adverse weather without per-frame ground truth, we further propose a proxy metric, Environment-Based KL Divergence (EBKL), under which 4D radar tensors yield more consistent predictions than camera and LiDAR in rain, sleet, and snow. The dataset and code will be publicly available.
\end{abstract}


\IEEEpeerreviewmaketitle

\section{INTRODUCTION}

\begin{figure*}[tb]
    \centering
    \includegraphics[width=1.85\columnwidth]{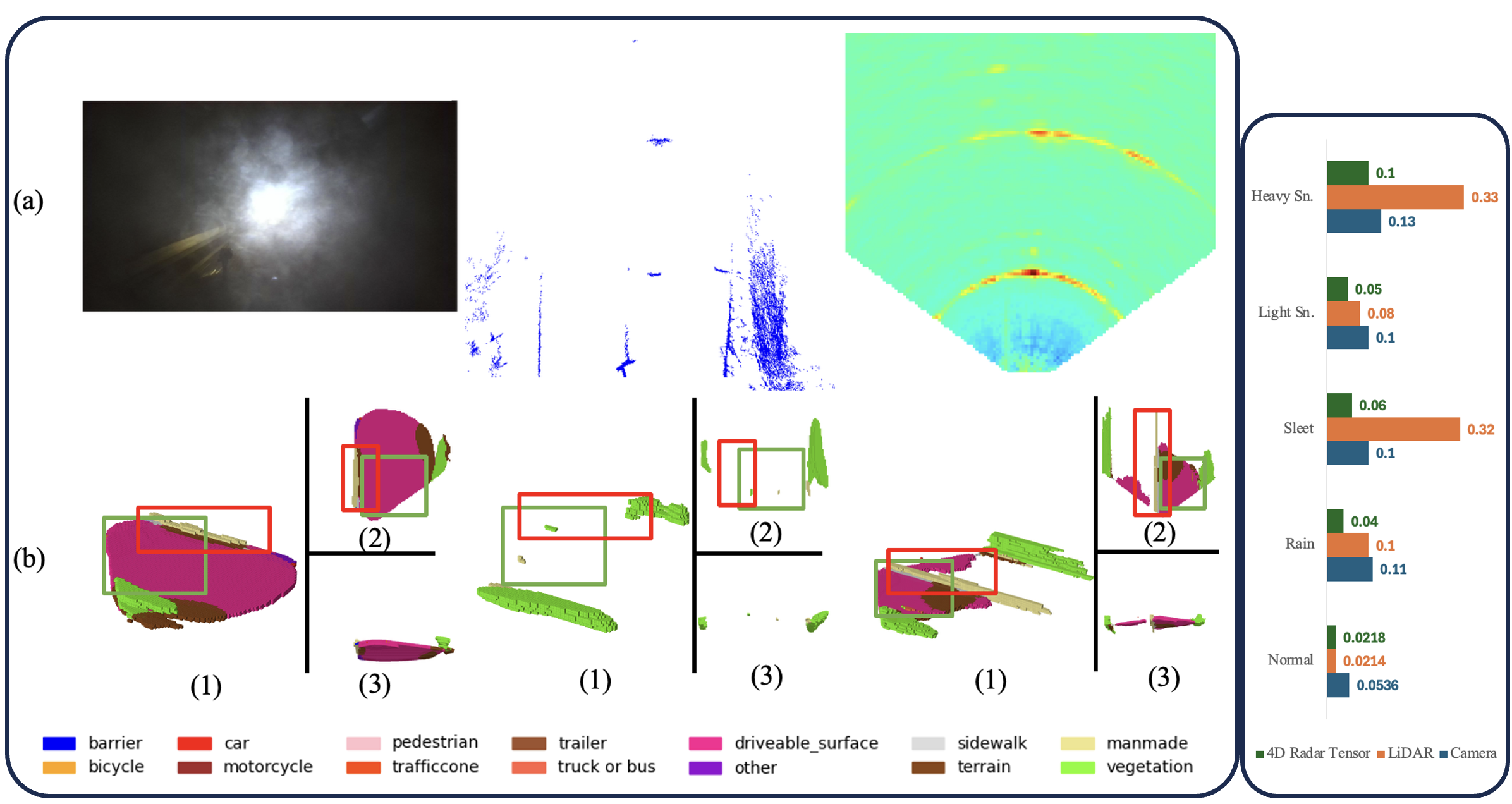}
    \caption{(Left) Visualization of 3D semantic occupancy prediction under heavy snow in a highway scenario.
    (a) Input modalities: (left) front-view camera image, (middle) LiDAR point cloud, and (right) BEV of 4D radar tensor.
    (b) Predicted occupancy using (left) camera-based OpenOcc-C~\cite{wang2023openoccupancylargescalebenchmark},
    (middle) LiDAR-based OpenOcc-L~\cite{wang2023openoccupancylargescalebenchmark}, and
    (right) the proposed RadarGSOcc with 4D radar input.
    RadarGSOcc produces more accurate predictions for key structures such as drivable areas (green box)
    and highway barriers (red box) under adverse weather. (Right) Quantitative comparison using the proposed EBKL proxy metric (lower is better).
    RadarGSOcc achieves lower EBKL, indicating more consistent predictions in adverse weather,
    while camera- and LiDAR-based methods degrade significantly.}
    \label{fig:adverse}
\end{figure*}

\IEEEPARstart{3}{D} semantic occupancy prediction enables autonomous vehicles to understand their 3D surroundings and identify safe navigable space for planning. By assigning an occupancy state and a semantic class to every voxel, it captures free space, drivable area, and objects of arbitrary shape that bounding-box detection cannot represent~\cite{tian2023occ3d, wang2023openoccupancylargescalebenchmark}. While camera- and LiDAR-based methods perform well under ideal conditions, they inherit the limitations of their sensors: cameras lose most of their information in low-light conditions, and LiDAR is severely degraded by rain, snow, and fog~\cite{wei2023surroundoccmulticamera3doccupancy, huang2024l4dr}.

4D radar provides a robust alternative. Millimeter-wave signals are largely insensitive to precipitation, fog, and illumination, and FMCW 4D radars measure the range, azimuth, elevation, and Doppler velocity of every reflection~\cite{kradar,abdu2021}. The raw 4D radar tensor (4DRT) retains the full power spectrum before any thresholding~\cite{kradar}, and its Doppler dimension directly encodes the radial velocity of each return, a strong cue for separating moving objects from static structures. Exploiting 4DRTs for dense prediction is nevertheless challenging: the tensor is large and noisy, so aggressive sparsification is required~\cite{ding2024radarocc}; its low angular resolution blurs fine structures; and the Doppler dimension mixes power and velocity information that standard backbones do not readily consume~\cite{Robust_3D_Objection_with_LiDAR_and_4d_Radar_Fusion}.

Radar-based occupancy prediction has only recently emerged. RadarOcc~\cite{ding2024radarocc} was the first to predict 3D occupancy directly from 4DRTs, and subsequent methods improve accuracy through LiDAR supervision~\cite{liu20254Drolls4Dradaroccupancy} or camera--radar fusion~\cite{yang2025metaoccsurroundview4Dradar, ninfa20264drc}, but require additional sensors. More importantly, existing radar-only methods underutilize Doppler cues: when we reproduce RadarOcc on our benchmark, its performance remains nearly unchanged with or without the Doppler input (Table~\ref{tab:doppler}).

In this paper, we propose \textbf{RadarGSOcc}, a Doppler-aware 3D semantic occupancy framework using 4D radar tensors. Our key idea is to extract the geometric (spatial power) and semantic (Doppler-related) information of the tensor separately and then fuse them. To this end, we introduce a \textbf{geometric-semantic feature extraction} module that processes the power and Doppler channels through separate sparse-convolution and deformable-attention branches~\cite{sparse_conv, xia2022visiontransformerdeformableattention} before fusing them, and a lightweight \textbf{geometric refinement} module in the occupancy head that is designed to recover fine-grained structures, such as small objects, using the geometric features.

We also construct the \textit{K-Radar 3DSemOcc} benchmark, the first large-scale benchmark for fine-grained 3D semantic occupancy prediction using 4D radar tensors, built from the K-Radar dataset~\cite{kradar}. It contains 14 semantic classes and an empty class over 5,910 training and 1,781 test frames. Labels are generated by a LiDAR-based occupancy model~\cite{wang2023openoccupancylargescalebenchmark} and refined with the 3D bounding-box annotations of K-Radar, which more than doubles the number of moving-object voxels.

On this benchmark, RadarGSOcc achieves \textbf{7.6\%} mIoU and \textbf{25.5\%} weighted mIoU within a 51.2\,m range, improving the weighted mIoU of RadarOcc by \textbf{16\%} relative (25.5 vs.\ 21.9) while using \textbf{11\% fewer parameters} and \textbf{5\% fewer FLOPs}; averaged over three training runs, the weighted mIoU is 23.8\% vs.\ 20.9\%. The gain comes mainly from large static classes such as drivable area, vegetation, and terrain. Vehicles remain difficult: RadarGSOcc improves truck/bus IoU but loses car IoU, and RadarOcc remains better when the two classes are merged into a single vehicle class. RadarGSOcc also outperforms camera-based~\cite{wang2023openoccupancylargescalebenchmark, li2024bevformer} and camera--radar fusion methods~\cite{wang2023openoccupancylargescalebenchmark, ninfa20264drc}, and narrows the gap with LiDAR-based approaches. Under adverse weather (e.g., rain, sleet, and snow), where camera and LiDAR degrade, its predictions remain more consistent. Since reliable ground truth is unavailable in such conditions, we propose a proxy metric, \textbf{Environment-Based KL Divergence (EBKL)}, which compares the predicted class distribution under a given weather condition with the normal-condition reference of the same environment. Under EBKL, 4D radar tensors give the lowest divergence of the three modalities in adverse weather.

In summary, the contributions of this paper are as follows:

\begin{itemize}
\item \textbf{K-Radar 3DSemOcc benchmark}: the first large-scale benchmark for fine-grained 3D semantic occupancy prediction using 4D radar tensors.
\item \textbf{RadarGSOcc}: an efficient Doppler-aware framework whose geometric-semantic feature extraction and geometric refinement modules improve radar-based occupancy prediction with fewer parameters and FLOPs than RadarOcc.
\item \textbf{Robustness and proxy metric}: a proxy metric (EBKL) for evaluating prediction consistency without ground truth, and an evaluation showing that 4D radar tensors give more consistent predictions than camera and LiDAR under adverse weather.
\end{itemize}

The remainder of this paper is organized as follows. Section~II reviews related work, Section~III presents the benchmark and method, Section~IV reports experiments, and Section~V concludes the paper. Supplementary material on the benchmark, implementation details, and additional results is appended after the main text.

\begin{table}[t]
\centering
\caption{Impact of Doppler index on RadarOcc performance (our reproduction on the K-Radar 3DSemOcc benchmark).}
\setlength{\tabcolsep}{6pt}
\renewcommand{\arraystretch}{1.1}
\begin{tabular}{l c c}
\hline
\textbf{Configuration} & \textbf{IoU} & \textbf{mIoU} \\
\hline \hline
RadarOcc (with Doppler) & 36.4 & 6.7 \\
RadarOcc (w/o Doppler)  & 35.9 ($-$0.5) & 6.7 (0.0) \\
\hline
\end{tabular}
\label{tab:doppler}
\end{table}

\section{RELATED WORK}

\subsection{4D Radar Datasets}

Large-scale camera and LiDAR datasets have significantly advanced autonomous driving research~\cite{semantic_kitti, klane}. In contrast, 4D radar datasets remain limited despite their robustness in adverse weather~\cite{ASTYX_3D_Radar_Dataset_for_Autonomous_Vehicle, zheng2022tj4Dradset, kradar, VOD_4D_Radar_Dataset_for_Autonomous_Vehicle}. Early datasets such as Astyx~\cite{ASTYX_3D_Radar_Dataset_for_Autonomous_Vehicle} contain only 500 frames, while VoD~\cite{VOD_4D_Radar_Dataset_for_Autonomous_Vehicle} and TJ4DRadSet~\cite{zheng2022tj4Dradset} mainly focus on urban scenes under clear conditions. K-Radar~\cite{kradar} addresses these limitations by providing 35,000 frames of 4D radar tensor data across diverse weather conditions, including rain, fog, and snow. However, it lacks annotations for 3D semantic occupancy prediction, motivating our proposed benchmark.

\subsection{3D Semantic Occupancy Prediction}

3D semantic occupancy prediction enables holistic scene understanding, including free space and drivable areas~\cite{goenawan2024unseen}. Occ3D~\cite{tian2023occ3d} introduced the first large-scale benchmark on nuScenes~\cite{caesar2020nuscenes}. Recent methods improve temporal modeling~\cite{wang2025unioccunifiedbenchmarkoccupancy}, point completion~\cite{EnhancingPerformanceOf3DPointCompletionNetwork}, multi-camera prediction~\cite{li2023fbocc3doccupancyprediction, wei2023surroundoccmulticamera3doccupancy}, and unified occupancy frameworks~\cite{wang2023openoccupancylargescalebenchmark}. Other works focus on efficiency and robustness through Gaussian-based occupancy~\cite{huang2025gaussianformer2,LoD-GS_Jiachen_Wang}, sparse prediction~\cite{liu2024fullysparse3doccprediction, wu2026sug}, label-noise handling~\cite{li2026trustunreliablevoxelsexploring}, and projection consistency~\cite{chambon2025gaussrender}.

\subsection{4D Radar-based Perception}

Early 4D radar research mainly focused on object detection. Methods such as RPFA-Net~\cite{RPFA-Net_4D_Radar_Pillar_Feature_Attention_Network} and SMURF~\cite{liu2023smurf} adapt LiDAR-based frameworks and reduce noise, while fusion-based approaches~\cite{xiong2023lxl, zheng2022tj4Dradset, peng2025mutualforcemutualawareenhancement4d} improve performance but are typically evaluated in clear-weather conditions. For direct radar tensor processing, RTNH and RTNH+~\cite{kradar, RTNH++Enhanced4DRadarObjectDetectionNetwork} learn radar-specific representations for 3D detection.

Recently, radar-based occupancy prediction has emerged. RadarOcc~\cite{ding2024radarocc} first applies 4D radar tensors to 3D occupancy prediction using transformer-based architectures~\cite{li2023voxformer, tian2023occ3d}. Subsequent works improve performance through LiDAR supervision~\cite{liu20254Drolls4Dradaroccupancy}, camera-radar fusion~\cite{yang2025metaoccsurroundview4Dradar}, and multi-modal learning~\cite{ninfa20264drc, zheng2026doracamom, LiCROcc_teach_radar_for_accurate_semantic_occupancy_prediction}.

Despite these advances, existing methods often struggle with accurate semantic prediction and rely on computationally expensive transformer architectures. In contrast, we propose an efficient framework that leverages geometric-semantic feature extraction and lightweight refinement for 3D semantic occupancy prediction from 4D radar tensors.

\section{METHOD}

\subsection{K-Radar 3DSemOcc Benchmark}
\label{sec:benchmark}

\begin{figure}[tb!]
\centering
\includegraphics[width=1.0\columnwidth]{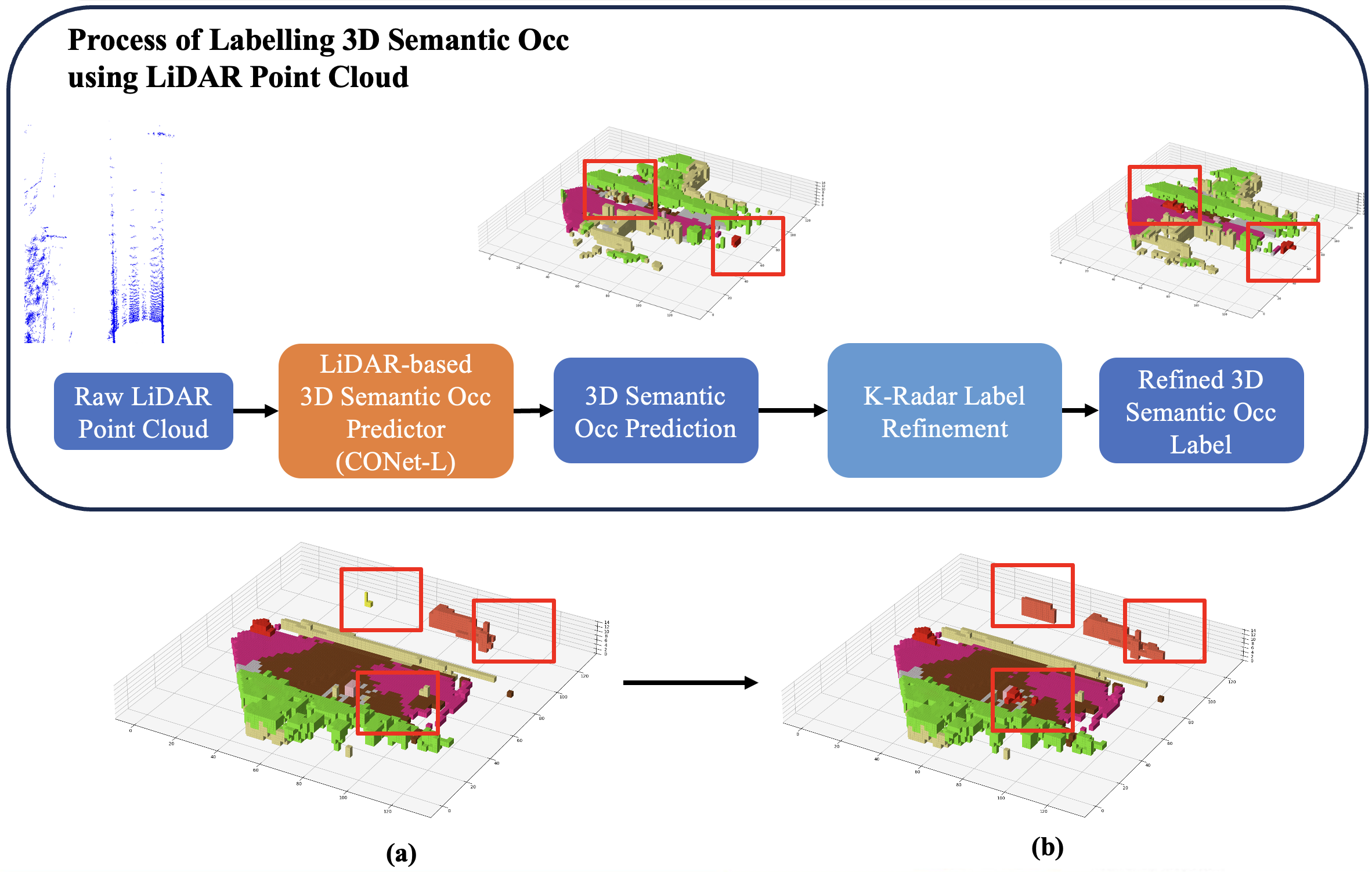}
\caption{Overview of the K-Radar 3DSemOcc labeling process. (a) Initial LiDAR-based labels. (b) Refined labels using K-Radar annotations, improving moving objects (red box).}
\label{fig:labeling}
\end{figure}

There is no large-scale dataset with fine-grained 3D semantic occupancy labels for 4D radar tensors: RadarOcc~\cite{ding2024radarocc} evaluates on K-Radar with occupancy labels that distinguish only foreground and background. To address this, we build a new benchmark based on K-Radar~\cite{kradar}. Following prior work~\cite{tian2023occ3d, wang2023openoccupancylargescalebenchmark}, we generate pseudo-labels from LiDAR using a single-frame strategy due to inaccurate odometry~\cite{RTNH++Enhanced4DRadarObjectDetectionNetwork}.

The labeling pipeline has two stages:

\textbf{(1) Initial labeling:} The LiDAR-based CONet-L model of OpenOccupancy~\cite{wang2023openoccupancylargescalebenchmark}, trained on nuScenes~\cite{caesar2020nuscenes}, generates 17-class pseudo-labels. We select scenes 2, 5, 8, 9, 10, 14, 16, 17, and 20 for training, and 3, 6, and 15 for testing; all of these scenes are recorded under normal weather conditions, and their environment, time of day, and number of frames are listed in Table~\ref{tab:supp_scenes} of the supplementary material.

\textbf{(2) Label refinement:} The pseudo-labels are refined with the 3D bounding-box annotations of K-Radar by assigning the object class (e.g., car, truck/bus, pedestrian) to the voxels within each bounding box. This more than doubles the number of moving-object voxels in the training split, from about 365k to 768k (Tables~\ref{tab:supp_class_before} and \ref{tab:supp_class_after}).

Fig.~\ref{fig:labeling} shows an example in which refinement recovers cars and trucks. We further simplify labels by merging bus and truck classes and removing construction vehicles. Manual verification of 525 randomly sampled non-empty voxels (threshold 0.8) gives an accuracy of \textbf{91.8\%} (482 correct); per-class results, including the moving-object classes, are reported in Table~\ref{tab:supp_verification} of the supplementary material together with the verification protocol. We retain voxels above a probability threshold, where 0.8 provides the best balance between quality and coverage (Section~\ref{sec:threshold}).

The final \textit{K-Radar 3DSemOcc} benchmark contains \textbf{5,910 training} and \textbf{1,781 testing frames} with 15 classes (14 semantic classes and the empty class), covering urban, university-campus, and highway environments under day and night conditions. The pipeline is general and can be extended to other datasets.

\subsection{RadarGSOcc: Doppler-Aware 3D Semantic Occupancy Prediction Using 4D Radar Tensor}

\begin{figure*}[tb!]
    \centering
    \includegraphics[width=1.8\columnwidth]{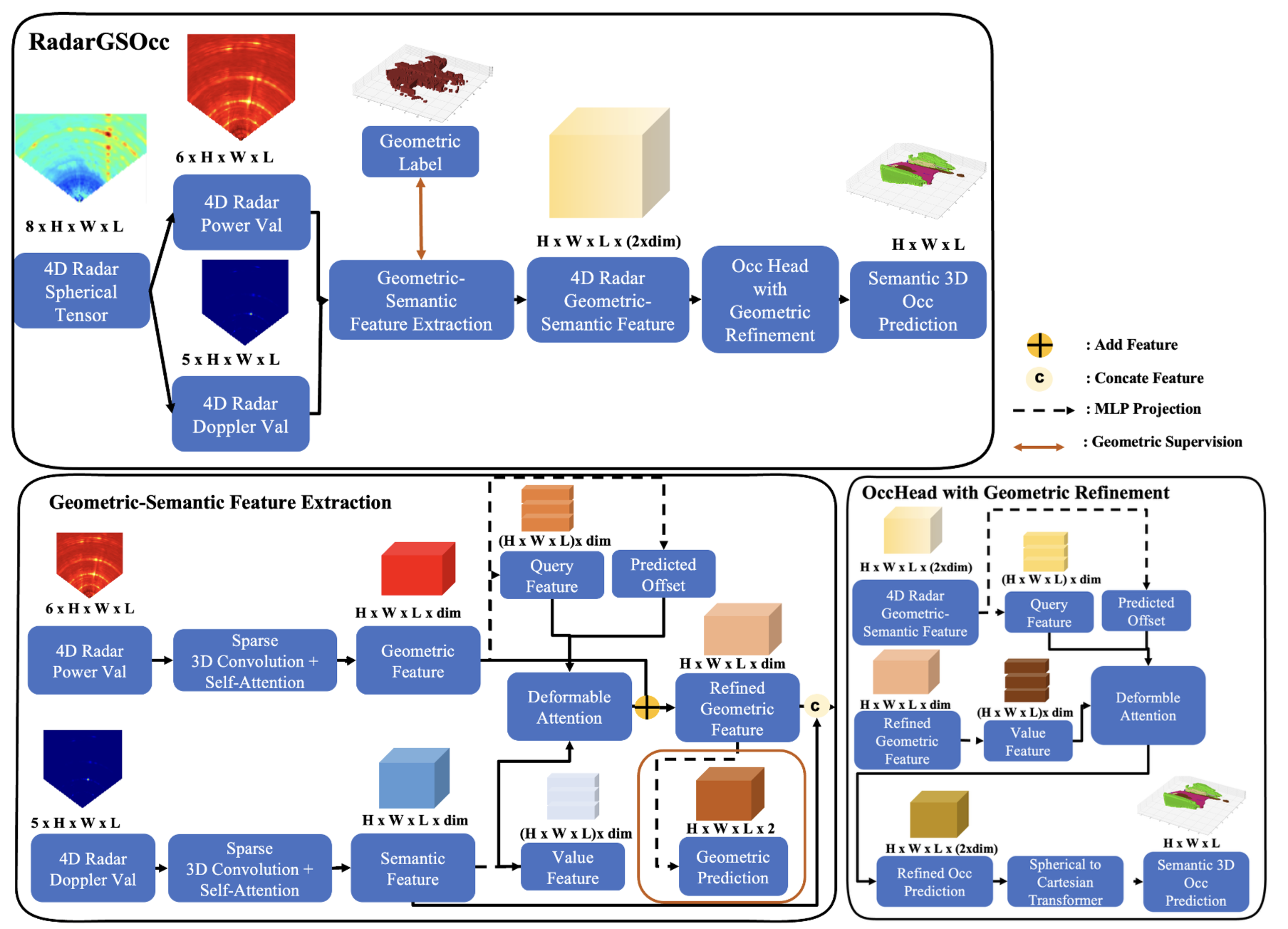}
    \caption{Overview of the proposed RadarGSOcc architecture. The input 4D radar tensor is divided into power and Doppler components and processed by the proposed Geometric-Semantic Feature Extraction module (bottom left, Section~\ref{sec:gsfe}), where $W \times H \times L$ denote spherical spatial dimensions and $dim$ is the feature dimension. The extracted features are then passed to the occupancy head (bottom right), where the proposed geometric feature refinement module (Section~\ref{sec:occhead}) refines occupancy features using geometric representations. The final output is a 3D semantic occupancy prediction of size $H \times W \times L$.}
    \label{fig:architecture}
\end{figure*}

\subsubsection{4D Radar Tensor Preprocessing}

The overall architecture of RadarGSOcc is shown in Fig.~\ref{fig:architecture}. We follow the preprocessing strategy of RadarOcc~\cite{ding2024radarocc}. Specifically, azimuth, range, and elevation are concatenated, and for each spatial element we extract the top-3 power values along the Doppler dimension together with their Doppler indices. To reduce computation, we keep only the top 250 values per range bin instead of using percentile-based sparsification~\cite{RTNH++Enhanced4DRadarObjectDetectionNetwork}. This produces a sparse 4D radar tensor that preserves key motion and structural information.

The final representation has 8 channels: the top-3 power values, the top-3 Doppler indices, and the mean and variance of the power along the Doppler dimension (Section~\ref{supp:doppler} of the supplementary material). This design follows RadarOcc while enabling efficient use of Doppler information for 3D semantic occupancy prediction.

\subsubsection{Geometric-Semantic Feature Extraction}
\label{sec:gsfe}

Extracting useful information from the Doppler dimension is challenging~\cite{Robust_3D_Objection_with_LiDAR_and_4d_Radar_Fusion}, since it contains both radar power and velocity-related information~\cite{kradar}. To address this, we design a feature extractor that separately learns geometric and semantic features from 4D radar tensors. Given the input tensor $x_{input} \in \mathbb{R}^{8 \times H_{voxel} \times W_{voxel} \times L_{voxel}}$, we apply sparse 3D convolutions~\cite{sparse_conv} to two channel groups:

\begin{equation}
x_{geo}= \text{Sparse3DConv}(x_{input}[:6,:,:,:]),
\end{equation}

\begin{equation}
x_{sem} = \text{Sparse3DConv}(x_{input}[3:,:,:,:]).
\end{equation}

Here, $x_{geo}$ captures geometric features, while $x_{sem}$ captures semantic and motion-related features. Channels 3--5 (the top-3 Doppler indices) are shared by both groups, since they are informative for both the geometric and the Doppler representations. Hence, the model can better exploit both power and Doppler information. Next, we enhance geometric features using a deformable self-attention mechanism inspired by \cite{xia2022visiontransformerdeformableattention}.
This deformable attention focuses only on selected regions, making it more efficient than full attention methods \cite{liu2021swintransformerhierarchicalvision, vit}. The 3D deformable self-attention yields the refined geometric feature $x'_{geo}$. The same process is applied to extract refined semantic features $x'_{sem}$. Implementation details of the deformable attention (number of sampling points and heads, offset normalization, and the interpolation scheme) are given in Section~\ref{supp:deformable} of the supplementary material.

\paragraph{Geometric-Semantic Fusion}

To further improve feature quality, we refine geometric features using semantic information via cross-attention, inspired by \cite{Robust_3D_Objection_with_LiDAR_and_4d_Radar_Fusion}. First, we project both features:
\begin{align}
& q_{fus} = MLP_q(x'_{geo}),\\
& k_{fus} = MLP_k(x'_{sem}),\\
& v_{fus} = MLP_v(x'_{sem}).
\end{align}

The normalized sampling offset $\Delta_{fus}$ is predicted as
\begin{equation}
\Delta_{fus} = MLP(x'_{geo}).
\end{equation}

We then apply deformable attention:
\begin{equation}
A_{geo\text{-}sem} = \text{Softmax}\left(\frac{q_{fus} \cdot k_{fus}[ref + \Delta_{fus}]^T}{\sqrt{dim}}\right),
\end{equation}

where $ref$ denotes the normalized coordinates of the reference points. The refined geometric feature in spherical coordinates is then
\begin{equation}
x_{geo-refined} = A_{geo-sem} \cdot v_{fus} + x'_{geo}.
\end{equation}

We also apply geometric supervision \cite{wei2023surroundoccmulticamera3doccupancy} by predicting whether each voxel contains an object:
\begin{equation}
x_{geo-supervised} = MLP(x_{geo-refined}).
\label{eq:geo_supervised}
\end{equation}

Finally, we combine geometric and semantic features:
\begin{equation}
x_{geo-sem} = \text{concat}[x_{geo-refined}, x'_{sem}].
\end{equation}

The fused features are then processed using a feature pyramid network (FPN) head \cite{fpn} and passed to the occupancy prediction head.

\subsubsection{OccHead with Geometric Feature Refinement}
\label{sec:occhead}

The extracted geometric-semantic features are processed by a multi-scale 3D convolutional occupancy head, following~\cite{wei2023surroundoccmulticamera3doccupancy}. However, similar to RadarOcc, the model tends to focus on large background regions (e.g., drivable area and terrain), while small objects (e.g., cars and trucks) are poorly captured due to loss of fine-grained features in earlier stages.

To address this issue, we introduce a \textbf{Geometric Feature Refinement} module based on 3D deformable attention~\cite{xia2022visiontransformerdeformableattention}, which is designed to enhance small-object representation using geometric features.

Let the occupancy feature be $x_{head} \in \mathbb{R}^{(H_{voxel} \times W_{voxel} \times L_{voxel}) \times dim}$ and the refined geometric feature be $x_{geo-refined}$, both in spherical coordinates. We project them into a shared latent space:
\begin{align}
& q_{ref} = MLP_{ref_q}(x_{head}),\\
& v_{ref} = MLP_{ref_v}(x_{geo-refined}).
\end{align}

Deformable attention is then computed as:
\begin{equation}
A_{ref} = \text{Softmax}\left(
\frac{q_{ref} \cdot v_{ref}[ref + \Delta_{ref}]^T}{\sqrt{dim}}
\right).
\end{equation}

The refined feature is obtained via residual fusion:
\begin{equation}
x_{head-refined} = MLP_{out}(A_{ref} \cdot v_{ref} + x_{head}).
\end{equation}

Following~\cite{ding2024radarocc}, features are transformed from spherical to Cartesian coordinates using a 3D deformable transformer. Finally, two 3D convolution layers with ReLU activation~\cite{Robust_3D_Objection_with_LiDAR_and_4d_Radar_Fusion} produce the output:
\begin{equation}
x_{pred} \in \mathbb{R}^{H_{voxel} \times W_{voxel} \times L_{voxel} \times N_{cls}},
\end{equation}
where $N_{cls}$ is the number of semantic classes.

\subsection{Loss Function}

The network is trained in a fully supervised manner using ground-truth occupancy labels. Following \cite{yang2025metaoccsurroundview4Dradar}, we adopt the cross-entropy loss $\textit{L}_{CE}$ as the primary objective. To address class imbalance, we additionally employ the Lovász-Softmax loss $\textit{L}_{lovasz\text{-}softmax}$ \cite{berman2018lovaszsoftmaxlosstractablesurrogate}. To further improve geometric and semantic consistency, we incorporate geometric and class-wise affinity losses, $\textit{L}_{geo}$ and $\textit{L}_{sem}$, as proposed in \cite{cao2022monoscenemonocular3dsemantic}. In addition, we apply a geometric supervision loss $\textit{L}_{geo\text{-}feat}$ on the radar geometric features (see~\eqref{eq:geo_supervised}) to enhance geometric feature learning; $\textit{L}_{geo\text{-}feat}$ is a binary cross-entropy loss whose target is the binarized ground-truth occupancy (occupied or empty) of each voxel. All loss terms are weighted equally, and the overall training objective is
\begin{equation}
\textit{L}_{overall} = \textit{L}_{CE} + \textit{L}_{lovasz\text{-}softmax} + \textit{L}_{geo} + \textit{L}_{sem} + \textit{L}_{geo\text{-}feat}.
\label{eq:loss}
\end{equation}

\section{EXPERIMENTS}

\subsection{Experiment Setup}
\label{sec:setup}

We compare RadarGSOcc with the camera-, LiDAR-, radar point cloud-, and camera--radar fusion baselines of OpenOccupancy~\cite{wang2023openoccupancylargescalebenchmark}, with BEVFormer~\cite{li2024bevformer} and 4DRC-Occ~\cite{ninfa20264drc}, and with the 4D radar tensor-based RadarOcc~\cite{ding2024radarocc}. For ablation, we evaluate RadarGSOcc variants without geometric refinement, without geometric-semantic fusion, and with standard 3D convolution refinement. For a fair comparison, RadarOcc and RadarGSOcc use the same feature dimension of 192. The prediction range is 0--51.2\,m forward, $\pm$25.6\,m laterally, and $-2.6$--$+3.0$\,m vertically, with a voxel grid of $128 \times 128 \times 14$ and 0.4\,m resolution. Models are trained on the 5,910 frames of the training scenes and evaluated on the 1,781 frames of the test scenes, covering diverse urban and highway environments under day and night conditions. All 4D radar point cloud-, camera-, and LiDAR-based methods follow their original training settings. OpenOcc-L (64) and OpenOcc-L (128) denote the LiDAR-based model with 64- and 128-channel LiDAR inputs, respectively. Since the official code of 4DRC-Occ is not publicly available, we use our own reproduction of its version-A model, implemented in the OpenOccupancy framework with the hyperparameters reported in the original paper; because the original results are reported on a different dataset, the reproduction could not be checked against them.

\subsection{Evaluation Metrics}
\label{sec:metrics}

We use binary IoU~\cite{ding2024radarocc}, mean IoU (mIoU)~\cite{sun2020scalabilityperceptionautonomousdriving}, and weighted mIoU (W-mIoU)~\cite{ninfa20264drc}, which weights each class by its voxel frequency so that extremely rare classes such as pedestrian and traffic cone (see Table~\ref{tab:supp_class_after} of the supplementary material) do not dominate the average:
\begin{equation}
\text{W-mIoU} = \sum_{c=1}^{C} \frac{n_c}{\sum_{i=1}^{C} n_i}\,\mathrm{IoU}_c,
\label{eq:wmiou}
\end{equation}
where $\mathrm{IoU}_c$ and $n_c$ denote the IoU and the number of ground-truth voxels of class $c$, respectively, and $C=14$ is the number of semantic classes.

Since accurate 3D semantic occupancy labels are difficult to obtain under adverse weather conditions due to degraded camera and LiDAR performance (Fig.~\ref{fig:adverse}), we introduce a supporting metric, \textbf{Environment-Based KL Divergence (EBKL)}:

\begin{equation}
x_{gt,env} = \text{Normalize} \left( \sum_{s=1}^{N_{scn}} x_{gt,s} \right),
\end{equation}

\begin{align}
&\text{EBKL}(x_{pred,env,weather}) \nonumber \\
&\quad = D_{\mathrm{KL}} \left( x_{gt,env} \,\|\, x_{pred,env,weather} \right) \nonumber \\
&\quad = \sum_{c \in \mathcal{C}_{E}}
x_{gt,env,c} \log \left(
\frac{x_{gt,env,c}}{x_{pred,env,weather,c}}
\right),
\label{eq:ebkl}
\end{align}

where $x_{pred,env,weather} \in \mathbb{R}^{|\mathcal{C}_{E}|}$ denotes the normalized predicted class distribution under a specific environment and weather condition, and $x_{gt,env} \in \mathbb{R}^{|\mathcal{C}_{E}|}$ denotes the normalized \emph{pseudo-label} class distribution aggregated from normal-condition scenes of the same environment. A small constant $\epsilon$ is added to $x_{pred,env,weather,c}$ before normalization to avoid divergence when a class is absent from the prediction. In the K-Radar dataset, environments include urban, university-campus, and highway scenarios. EBKL assumes that the class distribution remains relatively consistent within the same environment, allowing distributional consistency under adverse weather to be evaluated using the divergence from the normal-condition distribution. Therefore, we compute EBKL only over a class subset $\mathcal{C}_{E}$ consisting of relatively stable background classes (drivable area, terrain, manmade, barrier, and vegetation) and frequently occurring moving-object classes (car and truck); both distributions are renormalized over $\mathcal{C}_{E}$. This assumption is examined in Section~\ref{supp:ebkl} of the supplementary material, which shows that highway scenes are well separated from the other environments, whereas the urban and university-campus distributions are close to each other. The assumption may not hold across datasets with substantially different scene distributions, such as K-Radar versus nuScenes~\cite{caesar2020nuscenes} or Waymo~\cite{sun2020scalabilityperceptionautonomousdriving}; therefore, EBKL may be less reliable for cross-dataset comparisons. Since EBKL measures only the divergence between voxel-class distributions and does not evaluate voxel-level semantic accuracy, it cannot replace IoU-based metrics: in particular, a degenerate predictor that outputs the normal-condition class distribution regardless of its input would obtain EBKL~$=0$. We therefore use EBKL only as a \emph{proxy metric} for prediction consistency when ground-truth occupancy labels are unavailable under adverse weather, always together with the non-empty voxel ratio (Table~\ref{tab:weather}) and qualitative results (Fig.~\ref{fig:adverse}); its use is supported by its strong negative correlation with mIoU across models under normal conditions (Fig.~\ref{fig:miou_ebkl}).

\subsection{Implementation Details}
\label{sec:impl}

All 4D radar-based models are trained for 15 epochs on 4 RTX 3090 GPUs using PyTorch 2.0.0 and Ubuntu 18.04. We use a batch size of 1 and a learning rate of $3 \times 10^{-4}$. The training objective follows~\eqref{eq:loss}. Camera-based~\cite{wang2023openoccupancylargescalebenchmark, li2024bevformer}, fusion-based~\cite{wang2023openoccupancylargescalebenchmark, ninfa20264drc}, and LiDAR-based methods are trained with their original training settings.

\subsection{Experiment Results}
\label{sec:results}

\begin{table*}[t]
\centering
\caption{3D semantic occupancy results on the K-Radar 3DSemOcc benchmark (test split). Best, second-best, and third-best results are shown in bold, underlined, and italic, respectively. EBKL is computed under normal weather conditions (lower is better); ``--'' denotes not evaluated. ``+ w/o overlap'' denotes RadarGSOcc without the Doppler channels shared between the geometric and semantic branches. RadarOcc and RadarGSOcc rows are the first of three training runs (Section~\ref{sec:seeds}). Results for all classes are given in Table~\ref{tab:supp_full} of the supplementary material.}
\setlength{\tabcolsep}{3pt}
\renewcommand{\arraystretch}{1.1}
\begin{tabular}{l c | cccc | ccccccc}
\hline
\textbf{Method} & \textbf{Sensor}
& \textbf{IoU} & \textbf{mIoU} & \textbf{W-mIoU} & \textbf{EBKL}
& \textbf{DA} & \textbf{Veg} & \textbf{Man} & \textbf{Car} & \textbf{Truck}
& \textbf{Ter} & \textbf{Side} \\
\hline \hline

CONet-C\cite{wang2023openoccupancylargescalebenchmark} & Cam
& 16.9 & 3.2 & 11.3 & 0.093
& 27.4 & 4.2 & 0.0 & 0.6 & 0.0
& 11.4 & 0.0  \\

BEVFormer\cite{li2024bevformer} & Cam
& 27.3 & 4.1 & 16.0 & 0.072
& 34.0 & 10.4 & 0.1 & 0.0 & 0.0
& 11.9 & 0.5  \\

OpenOcc-C\cite{wang2023openoccupancylargescalebenchmark} & Cam
& 30.0 & 5.4 & 17.1 & 0.054
& 34.7 & 17.3 & \textit{13.7} & \underline{5.9} & 1.8
& 19.0 & 0.9  \\

OpenOcc-R\cite{wang2023openoccupancylargescalebenchmark} & Radar
& 26.0 & 4.0 & 14.7 & 0.054
& 34.0 & 8.8 & 5.1 & 1.2 & 0.0
& 8.9 & 0.8  \\

OpenOcc-RC\cite{wang2023openoccupancylargescalebenchmark} & R+Cam
& 25.2 & 3.8 & 14.0 & 0.075
& 34.9 & 5.6 & 0.8 & 0.0 & 0.1
& 11.0 & 1.4  \\

4DRC-Occ\cite{ninfa20264drc} & R+Cam
& 30.2 & 4.6 & 17.6 & 0.054
& 35.2 & 12.8 & 0.1 & 0.2 & 0.0
& 15.3 & 1.9  \\

\hline

OpenOcc-L (64)\cite{wang2023openoccupancylargescalebenchmark} & LiDAR
& \underline{53.1} & \underline{10.9} & \underline{36.1} & \textbf{0.021}
& \underline{50.5} & \underline{39.9} & \underline{15.4} & \textit{3.3} & \textit{2.4}
& \underline{26.1} & \textbf{9.1}  \\

OpenOcc-L (128)\cite{wang2023openoccupancylargescalebenchmark} & LiDAR
& \textbf{54.3} & \textbf{11.1} & \textbf{38.9} & --
& \textbf{51.5} & \textbf{43.0} & \textbf{15.4} & 1.5 & \underline{3.3}
& \textbf{30.3} & \underline{6.4}  \\

RadarOcc\cite{ding2024radarocc} & RadarT
& 36.4 & 6.7 & 21.9 & \textit{0.032}
& 40.4 & 16.3 & 10.3 & \textbf{6.3} & 0.6
& 18.0 & 2.4  \\

\hline

\textbf{RadarGSOcc(Ours)} & RadarT
& \textit{38.2} & \textit{7.6} & \textit{25.5} & \underline{0.022}
& \textit{42.2} & \textit{21.2} & 11.6 & 0.6 & \textbf{4.7}
& \textit{23.9} & 2.5  \\

+ w/o overlap & RadarT
& 35.7 & 6.8 & 21.8 & --
& 34.5 & 17.5 & 11.9 & 0.6 & 2.1
& 21.2 & \textit{4.9}  \\

\hline
\end{tabular}
\label{tab:main}
\end{table*}

\subsubsection{Result of Proposed RadarGSOcc against State-of-the-Art Methods}
\label{sec:sota}

\begin{figure*}[tb!]
\centering
\includegraphics[width=1.8\columnwidth]{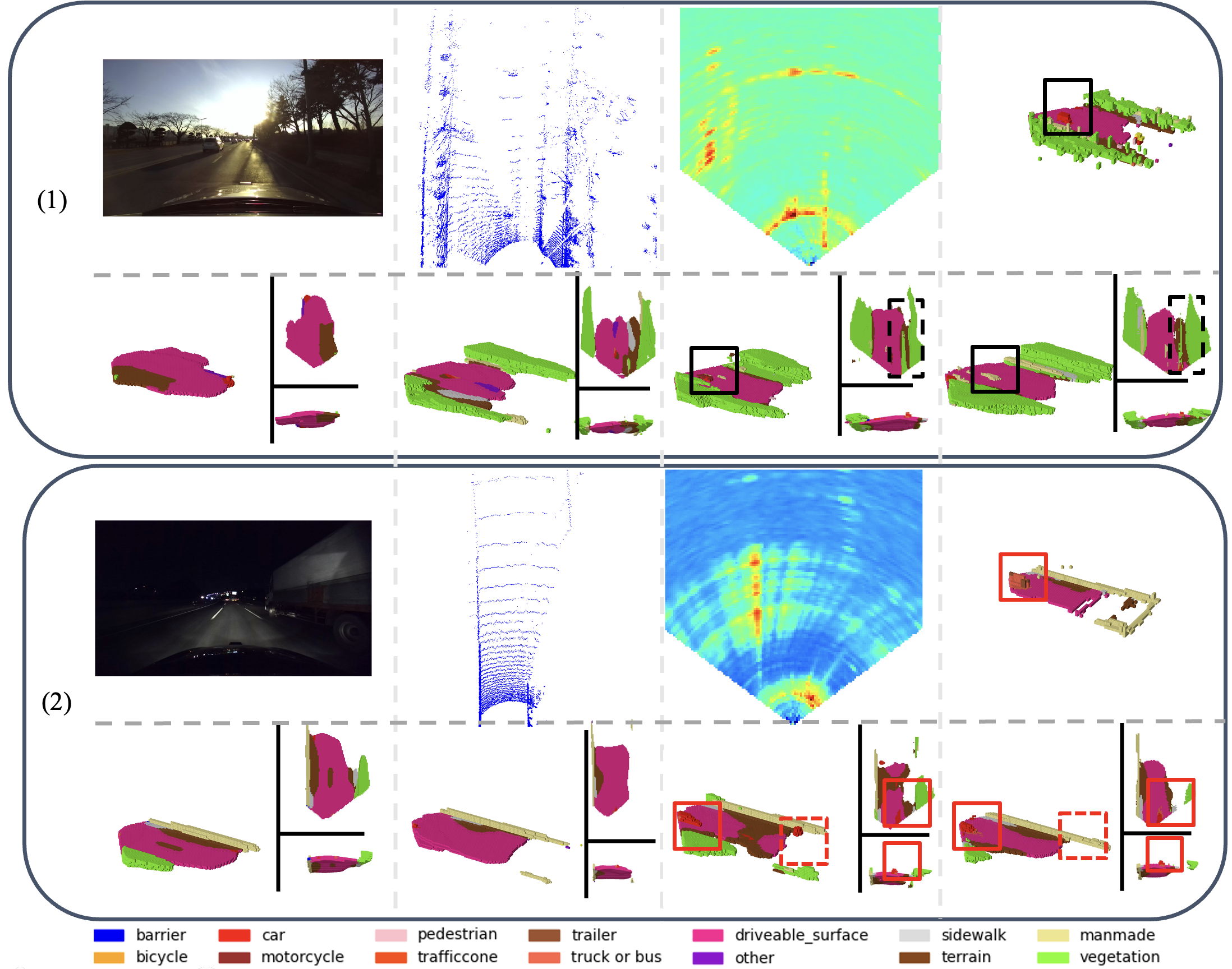}
\caption{Visualization of 3D semantic occupancy prediction using LiDAR (OpenOcc-L)~\cite{wang2023openoccupancylargescalebenchmark}, camera (OpenOcc-C)~\cite{wang2023openoccupancylargescalebenchmark}, and 4D radar tensors with RadarOcc~\cite{ding2024radarocc} and the proposed RadarGSOcc in urban (1) and highway (2) scenarios. In each scenario, the top row shows (from left) the front-camera image, LiDAR point cloud, 4D radar tensor, and ground truth, while the bottom row shows camera-based prediction, LiDAR-based prediction, RadarOcc, and RadarGSOcc. Each prediction includes a 3D front-right view, BEV view, and front view. In these examples, RadarGSOcc predicts the highlighted moving objects (red and black boxes) more accurately, improves terrain and drivable area prediction (dashed black box), and reduces false moving-object predictions (dashed red box); quantitatively, however, its merged vehicle IoU is lower than that of RadarOcc (Section~\ref{sec:sota}).}
\label{fig:qualitative}
\end{figure*}

As shown in Table~\ref{tab:main}, RadarGSOcc reaches 25.5\% W-mIoU, a 16\% relative improvement over RadarOcc~\cite{ding2024radarocc} (21.9\%), with about 11\% fewer parameters and 5\% fewer FLOPs (Table~\ref{tab:ablation}), although its frame rate is slightly lower (3.1 vs.\ 3.3 FPS). The improvement comes mainly from the large static classes that dominate the voxel-weighted average: drivable area (42.2\% vs.\ 40.4\%), vegetation (21.2\% vs.\ 16.3\%), and terrain (23.9\% vs.\ 18.0\%). For vehicles, the result is mixed. RadarGSOcc attains the highest truck/bus IoU of all methods (4.7\%), but its car IoU drops from 6.3\% to 0.6\%, and when car and truck/bus are merged into a single vehicle class, RadarOcc remains better than RadarGSOcc (6.5\% vs.\ 2.6\% IoU). The higher truck/bus IoU therefore does not translate into better vehicle occupancy overall, and vehicle prediction from 4D radar tensors remains an open problem. Removing the channel overlap reduces W-mIoU by 3.7 points (25.5\% $\rightarrow$ 21.8\%) and truck/bus IoU from 4.7\% to 2.1\%, while recovering some rare small-object classes (e.g., trailer and pedestrian; Table~\ref{tab:supp_full}), indicating that the shared Doppler-index channels mainly benefit the dominant classes.

RadarGSOcc also outperforms the camera-based and camera--radar fusion baselines~\cite{wang2023openoccupancylargescalebenchmark, li2024bevformer, ninfa20264drc} in IoU, mIoU, and W-mIoU. Qualitative examples are shown in Fig.~\ref{fig:qualitative}.

All methods still struggle on small foreground classes such as cones, pedestrians, and motorcycles because these classes each account for less than 0.1\% of voxels (Table~\ref{tab:supp_class_after}), leading to severe class imbalance.

\subsubsection{Training with Different Random Seeds}
\label{sec:seeds}
We train RadarGSOcc and the baseline RadarOcc~\cite{ding2024radarocc} three times with different random seeds to evaluate training robustness (Table~\ref{tab:supp_seeds} of the supplementary material). Averaged over the three runs, RadarGSOcc achieves 37.43\% $\pm$ 2.34\% IoU, 7.14\% $\pm$ 0.41\% mIoU, and 23.83\% $\pm$ 2.15\% W-mIoU, compared with 36.26\% $\pm$ 0.23\%, 6.72\% $\pm$ 0.28\%, and 20.93\% $\pm$ 0.90\% for RadarOcc. RadarGSOcc is therefore better on average, but not in every run: in the third run, its IoU and mIoU (34.8\% and 6.8\%) are lower than those of RadarOcc (36.0\% and 7.0\%), although its W-mIoU remains higher (21.4\% vs.\ 20.8\%). RadarGSOcc also shows a much larger run-to-run variation (e.g., $\pm$2.15 vs.\ $\pm$0.90 in W-mIoU), so its training is less stable than that of RadarOcc. Table~\ref{tab:main} and the other tables report the first run of each model, which is also the best of the three runs in W-mIoU for both models.

\subsubsection{Effect of Class Probability Threshold}
\label{sec:threshold}

\begin{table}
    \centering
    \caption{Performance of RadarGSOcc on the K-Radar 3DSemOcc benchmark under different class probability thresholds.
    }
    \begin{tabular}{c|c|c|c|c}
         \hline
         \textbf{Threshold} & \textbf{mIoU} & \textbf{W-mIoU} & \textbf{\#Voxel GT} & \textbf{\#Voxel Prediction} \\
         \hline
         \hline
         0.2 & 3.97 & 12.7 & 26,647 & 14,842\\
         0.4 & 5.44 & 17.7 & 25,796 & 17,130 \\
         0.8 & 7.60 & 25.5 & 14,606 & 9,930 \\
         0.9 & 7.14 & 24.5 & 10,559 & 7,049 \\
         \hline
    \end{tabular}
    \label{tab:threshold}
\end{table}

We evaluate different label-confidence thresholds (Table~\ref{tab:threshold}). Since the threshold defines the evaluation label set itself, scores across rows are not directly comparable and only characterize the quality--coverage trade-off: lower thresholds keep noisier pseudo-labels, higher ones discard valid voxels. At 0.8, the manual verification gives a label accuracy of 91.8\% (Section~\ref{sec:benchmark}); we adopt this threshold for all main experiments.

\subsubsection{Result of Proposed Geometric-Semantic Feature Extractor and Geometric Refinement}
\label{sec:ablation}

\begin{table}[t]
\centering
\caption{Ablation study of RadarGSOcc. Fusion: geometric-semantic fusion; Conv/Def: geometric refinement with 3D convolution / deformable attention. $\checkmark$ indicates enabled. Param in millions (M) and FLOPs in GFLOPs (G). Each variant is a single training run.}
\setlength{\tabcolsep}{4pt}
\renewcommand{\arraystretch}{1.05}
\begin{tabular}{ccc|ccc|ccc}
\hline
\textbf{Fusion} & \textbf{Conv} & \textbf{Def}
& \textbf{IoU} & \textbf{mIoU} & \textbf{W-mIoU}
& \textbf{Param} & \textbf{FLOPs} & \textbf{FPS} \\
\hline \hline

\multicolumn{3}{c|}{RadarOcc~\cite{ding2024radarocc}}
& 36.4 & 6.7 & 21.9
& 117.5 & 1334.1 & \textbf{3.3} \\

\hline

$\times$ & $\times$ & $\times$
& 35.7 & 6.9 & 21.8
& \textbf{104.2} & \textbf{1260.1} & 3.2 \\

$\times$ & $\times$ & $\checkmark$
& 36.1 & 7.0 & 21.1
& 104.6 & 1270.8 & 3.1 \\

$\checkmark$ & $\times$ & $\times$
& 35.9 & 6.6 & 20.7
& \textbf{104.2} & 1262.7 & 3.1 \\

$\checkmark$ & $\checkmark$ & $\times$
& 32.3 & 5.6 & 18.1
& 105.0 & 1285.4 & 3.2 \\

$\checkmark$ & $\times$ & $\checkmark$
& \textbf{38.2} & \textbf{7.6} & \textbf{25.5}
& 104.6 & 1273.5 & 3.1 \\

\hline
\end{tabular}
\label{tab:ablation}
\end{table}

\begin{figure}[tb!]
\centering
\includegraphics[width=1.0\columnwidth]{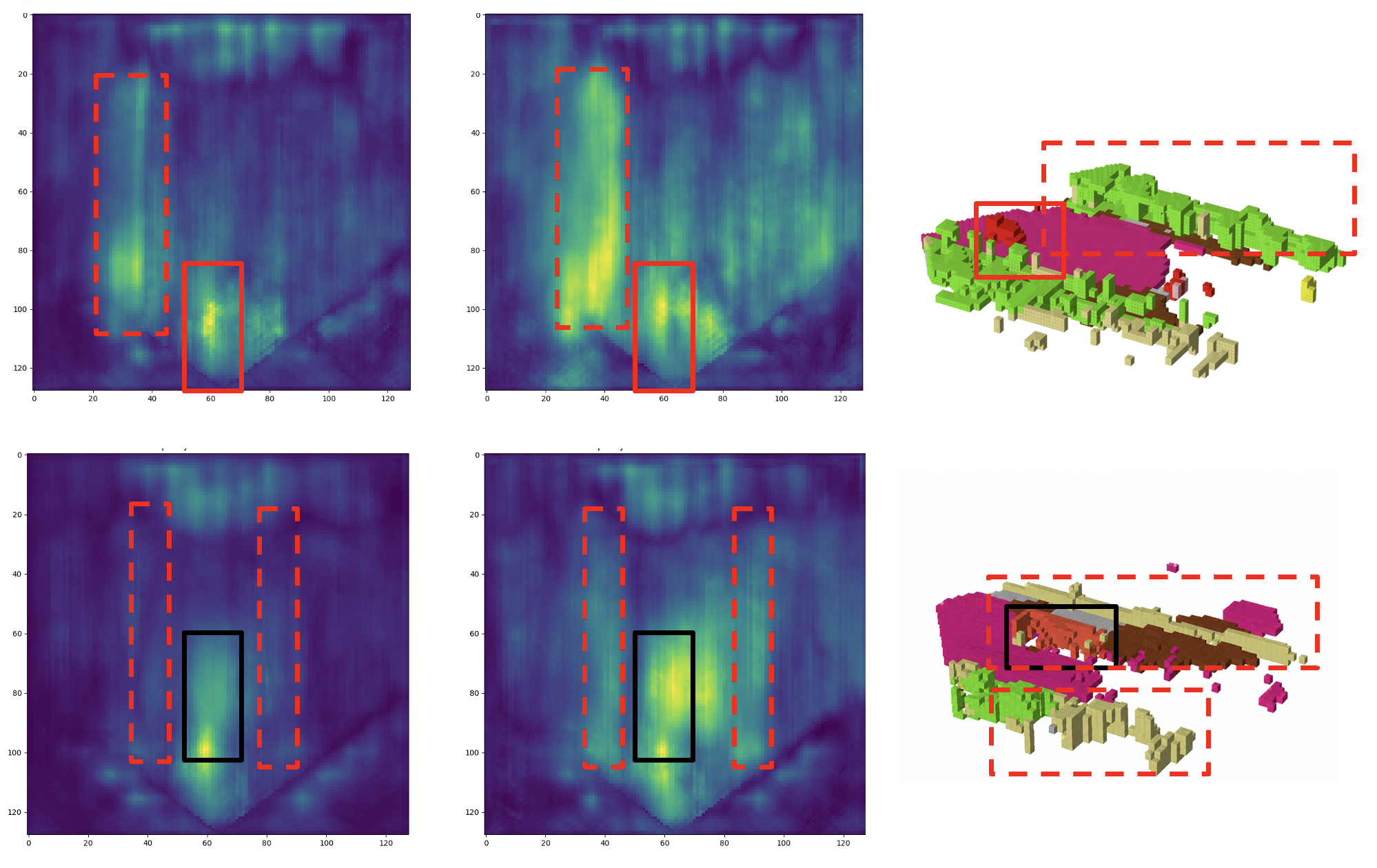}
\caption{Visualization of 3D occupancy features before (left) and after (middle) geometric refinement, compared with ground truth (right). In these examples, the refined features respond more strongly to moving objects (cars and trucks in red and black boxes) and road structures (dashed red boxes).}
\label{fig:refinement}
\end{figure}

We conduct ablation studies to evaluate geometric-semantic fusion and geometric refinement (Table~\ref{tab:ablation}), in which RadarOcc processes all input channels jointly and applies no refinement in the dense head. Separating the geometric and semantic branches alone (first row) already matches RadarOcc (6.9\% vs.\ 6.7\% mIoU) with fewer parameters and FLOPs. Neither module helps on its own: the deformable geometric refinement without fusion changes W-mIoU only marginally (21.8\% $\rightarrow$ 21.1\%), and geometric-semantic fusion without refinement slightly degrades it (20.7\%), while refining the fused features with standard 3D convolution degrades it further (18.1\%), indicating that a local convolutional refinement cannot exploit the fused geometric features. Only the combination of fusion and deformable-attention refinement yields a clear gain, raising W-mIoU to 25.5\% (+3.7 over the base model and +7.4 over convolutional refinement; 23.8\% on average over three runs, Section~\ref{sec:seeds}) at a cost of only +0.4M parameters and +13 GFLOPs. The two modules therefore appear complementary: fusion injects Doppler-informed cues into the geometric features, and the deformable refinement allows the occupancy head to exploit them. Because each variant is trained once and the run-to-run standard deviation of RadarGSOcc is about 2 points in W-mIoU (Section~\ref{sec:seeds}), differences of about one point among the other variants should not be over-interpreted. The geometric refinement module slightly lowers the frame rate (3.2 $\rightarrow$ 3.1 FPS) but improves W-mIoU. In the examples of Fig.~\ref{fig:refinement}, the refined features respond more strongly to moving objects (cars and trucks) and road structures.

\subsubsection{Performance with Various Radar Tensor Processing and Top Doppler Dimension}
\label{sec:topk}

\begin{table}[t]
\centering
\caption{Ablation of RadarGSOcc with different Top-K values and Doppler dimensions (single training runs).}
\setlength{\tabcolsep}{3pt}
\renewcommand{\arraystretch}{1.05}
\begin{tabular}{c c | ccc | cc}
\hline
\textbf{Top-K} & \textbf{Doppler}
& \textbf{IoU} & \textbf{mIoU} & \textbf{W-mIoU}
& \textbf{FLOPs} & \textbf{FPS} \\
\hline \hline

150 & 1 & 35.2 & 6.3 & 20.0 & \textbf{1273.2} & \textbf{3.2} \\
150 & 2 & 35.8 & 6.8 & 21.2 & \textbf{1273.2} & \textbf{3.2} \\
150 & 3 & 36.5 & 6.8 & 20.4 & \textbf{1273.2} & \textbf{3.2} \\

\hline

250 & 1 & 36.9 & 6.6 & 21.5 & 1273.3 & 3.1 \\
250 & 2 & 34.5 & 6.8 & 20.2 & 1273.4 & 3.1 \\
250 & 3 & \textbf{38.2} & \textbf{7.6} & \textbf{25.5} & 1273.5 & 3.1 \\

\hline
\end{tabular}
\label{tab:topk}
\end{table}

We evaluate different Top-K values and Doppler dimensions (Table~\ref{tab:topk}). The best
configuration uses Top-K = 250 with all three Doppler
components, reaching 25.5\% W-mIoU. The effect of the
Doppler dimension is not strictly monotonic at either Top-K, but the third Doppler component is clearly beneficial only at the higher Top-K = 250 (20.2\% $\rightarrow$ 25.5\%), where denser measurements make the additional component informative; at Top-K = 150 it instead adds noise (21.2\% $\rightarrow$ 20.4\%).
This indicates that the benefit of finer Doppler resolution depends on sufficient measurement density. Across these Top-K and Doppler settings, FLOPs and frame rate remain nearly unchanged (Table~\ref{tab:topk}), as they affect only the sparse input representation, not the network structure.

\subsubsection{Performance in Adverse Weather Conditions}
\label{sec:adverse}

\begin{figure}[tb!]
\centering
\includegraphics[width=0.95\columnwidth]{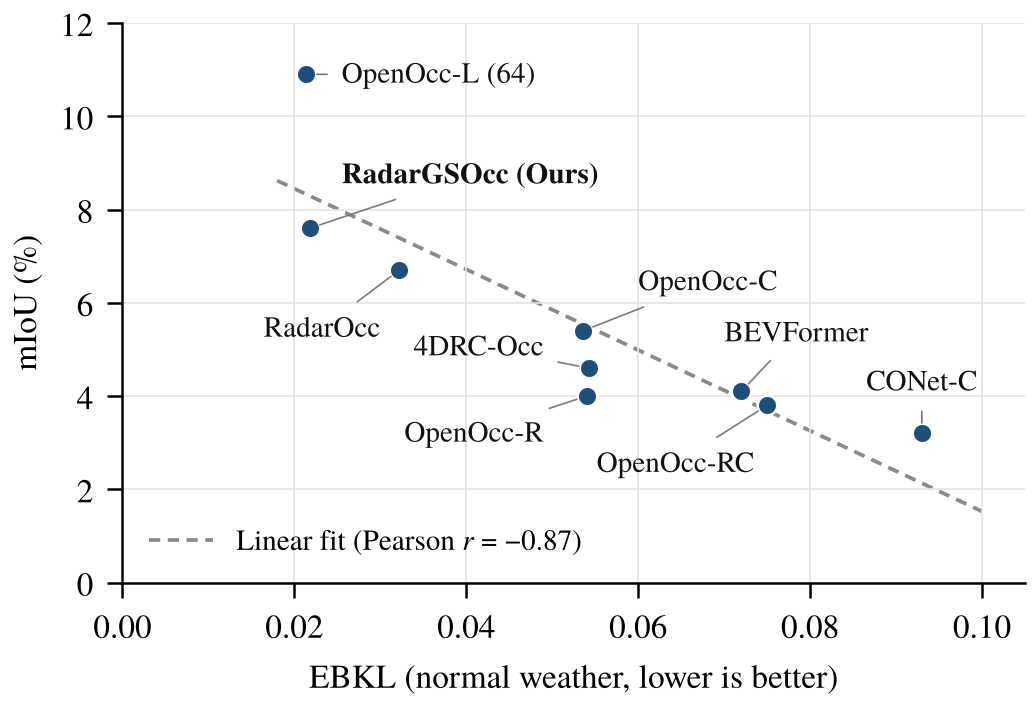}
\caption{mIoU versus EBKL (normal weather, test split) of the nine models in Table~\ref{tab:main} for which EBKL is evaluated. The dashed line is a least-squares fit; the values are listed in Table~\ref{tab:supp_corr} of the supplementary material.}
\label{fig:miou_ebkl}
\end{figure}

\begin{table}[t]
\centering
\caption{Percentage of non-empty voxel predictions under different weather conditions (Light/Heavy: light/heavy snow). The adverse-weather sequences are listed in Table~\ref{tab:supp_scenes} of the supplementary material.}
\setlength{\tabcolsep}{4pt}
\renewcommand{\arraystretch}{1.1}
\begin{tabular}{l c ccccc}
\hline
\textbf{Model} & \textbf{Normal} & \textbf{Sleet} & \textbf{Rain} & \textbf{Fog} & \textbf{Light} & \textbf{Heavy} \\
\hline \hline
OpenOcc-C\cite{wang2023openoccupancylargescalebenchmark} & 18.5 & 17.5 & 16.7 & 16.5 & 17.0 & 18.9 \\
CONet-L\cite{wang2023openoccupancylargescalebenchmark}   & 29.8 & 9.7 & 14.3 & 9.5 & 9.8 & 14.2 \\
RadarGSOcc & 26.5 & 26.1 & 24.5 & 28.1 & 15.1 & 15.3 \\
\hline
\end{tabular}
\label{tab:weather}
\end{table}

We evaluate RadarGSOcc under adverse weather conditions on the unlabeled adverse-weather sequences of the K-Radar dataset (Fig.~\ref{fig:adverse} and Table~\ref{tab:supp_scenes}). Since reliable per-frame ground truth is unavailable, we use the non-empty voxel ratio (how much occupancy a
model predicts) and the proposed EBKL (how close its class distribution stays to the normal-condition reference) as complementary proxies. As shown in Fig.~\ref{fig:adverse} (right), RadarGSOcc attains lower EBKL than the camera- and LiDAR-based methods in rain, sleet, light snow, and heavy snow. Across the nine models evaluated under normal weather, lower EBKL goes with higher mIoU (Fig.~\ref{fig:miou_ebkl}; Pearson correlation coefficient $-0.87$), indicating that EBKL tracks mIoU. Table~\ref{tab:weather} is consistent with this: LiDAR collapses under sleet, fog, and snow (e.g., 29.8\% $\rightarrow$ 9.5\% in fog), the camera model is stable but low, and RadarGSOcc preserves most of its occupancy through sleet, rain, and fog. In snow, however, the voxel ratio of RadarGSOcc drops to about 15\% in both light and heavy snow, and under heavy snow its EBKL also doubles relative to light snow (0.05 $\rightarrow$ 0.10), which we attribute to increased 4D radar noise in dense snow~\cite{kradar}. Because the adverse-weather sequences also differ from the normal-weather test scenes in environment and time of day (Table~\ref{tab:supp_scenes}), both proxies reflect scene changes as well as weather. Overall, 4D radar tensors give the most consistent predictions of the three modalities, with snow as the main failure case~\cite{Robust_3D_Objection_with_LiDAR_and_4d_Radar_Fusion}.

\section{CONCLUSION AND FUTURE WORK}

In this work, we introduce the \textit{K-Radar 3DSemOcc} benchmark, the first large-scale benchmark for fine-grained 3D semantic occupancy prediction using 4D radar tensors. Built on the K-Radar dataset~\cite{kradar}, it is generated using LiDAR-based occupancy prediction with refinement from radar annotations, covering diverse urban and highway scenes under day and night conditions. We also propose RadarGSOcc, an efficient framework that leverages geometric-semantic feature extraction and geometric refinement for 3D semantic occupancy prediction from 4D radar tensors. RadarGSOcc achieves \textbf{25.5\%} weighted mIoU and \textbf{7.6\%} mIoU, a \textbf{16\%} relative improvement over RadarOcc~\cite{ding2024radarocc} in weighted mIoU, while using \textbf{11\% fewer parameters} and 5\% fewer FLOPs. It also outperforms camera-based~\cite{wang2023openoccupancylargescalebenchmark, li2024bevformer} and camera--radar fusion methods, and gives more consistent predictions than camera and LiDAR under adverse weather such as rain, sleet, and snow. Its main limitations are vehicle prediction, where RadarOcc remains better when car and truck/bus are merged, and a larger run-to-run variation than RadarOcc.

Furthermore, we propose \textbf{Environment-Based KL Divergence (EBKL)} as a proxy metric to evaluate prediction consistency under adverse weather without ground truth. RadarGSOcc achieves better results in both EBKL and qualitative visualization, highlighting the effectiveness of 4D radar for robust 3D semantic occupancy prediction in adverse weather.

For future work, we aim to improve vehicle prediction and training stability, improve label quality using larger-scale datasets with accurate human annotations, and further advance robust 4D radar-based occupancy prediction in real-world environments.

\section*{ACKNOWLEDGMENT}

This work was supported by the Ministry of Trade, Industry and Energy (MOTIE) grant funded by the Korea government (No. RS-2025-25451359).

{\small
\bibliographystyle{IEEEtran}
\bibliography{main}
}

\begin{IEEEbiography}[{\includegraphics[width=1in,height=1.25in,clip,keepaspectratio]{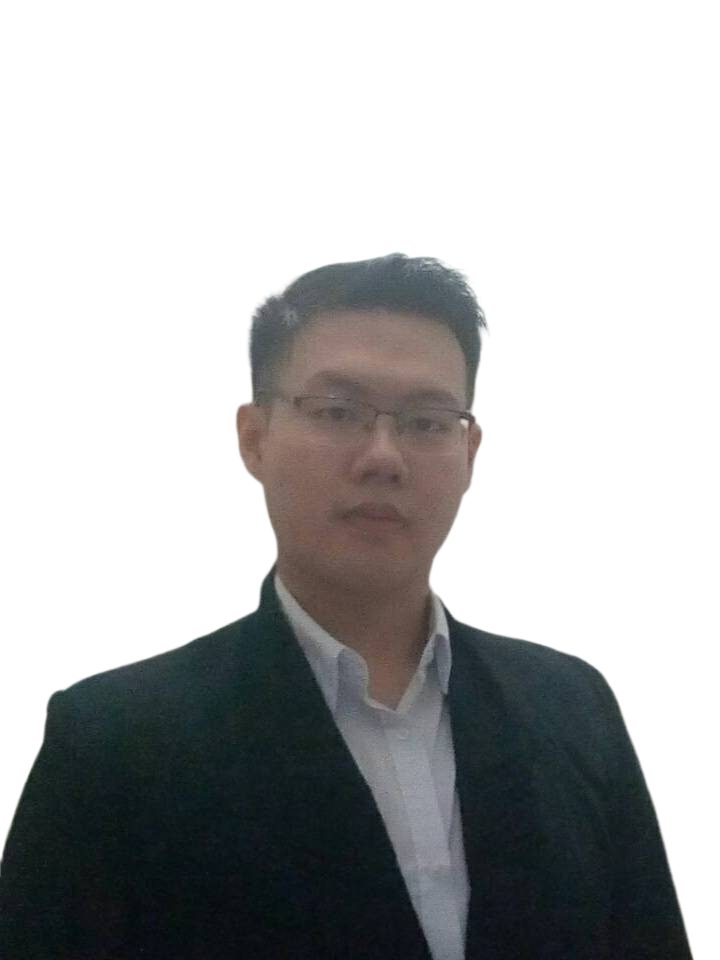}}]{Christofel Rio Goenawan}
received the B.S. degree in electrical engineering from the Bandung Institute of Technology (ITB), Indonesia, in 2021. He is currently pursuing the M.S. degree with the Robotics Program, Korea Advanced Institute of Science and Technology (KAIST), Daejeon, South Korea. He has three years of industry experience as a data scientist and machine learning engineer in the telecommunication, FMCG, banking, and marketing sectors. His current research interests include perception and end-to-end autonomous driving using AI.
\end{IEEEbiography}

\begin{IEEEbiography}[{\includegraphics[width=1in,height=1.25in,clip,keepaspectratio]{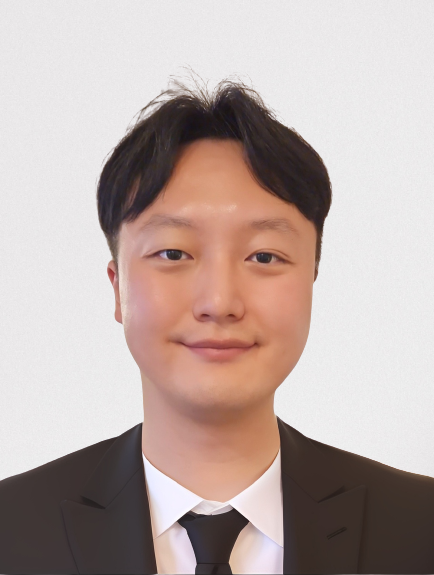}}]{Dong-Hee Paek}
(Member, IEEE) is currently an Assistant Professor with the Department of Future Mobility, Korea University, Sejong, Korea. He received the B.S. degree in robotics and intelligent systems from Kwangwoon University, Seoul, Korea, in 2019, and the M.S. degree in robotics and the Ph.D. degree in mobility and robotics from the Korea Advanced Institute of Science and Technology (KAIST), Daejeon, Korea, in 2021 and 2025, respectively. From 2024 to 2025, he was with Zeta Mobility, Daejeon, Korea, as a Senior Research Engineer, where he conducted research and development on 4D radar perception and sensor fusion AI while pursuing the Ph.D. degree. From 2025 to 2026, he was a Postdoctoral Research Fellow with KAIST. His current research interests include 4D radar signal processing and perception, multi-modal sensor fusion, and AI-based robotic perception.
\end{IEEEbiography}

\begin{IEEEbiography}
[{\includegraphics[width=1in,height=1.25in,clip,keepaspectratio]{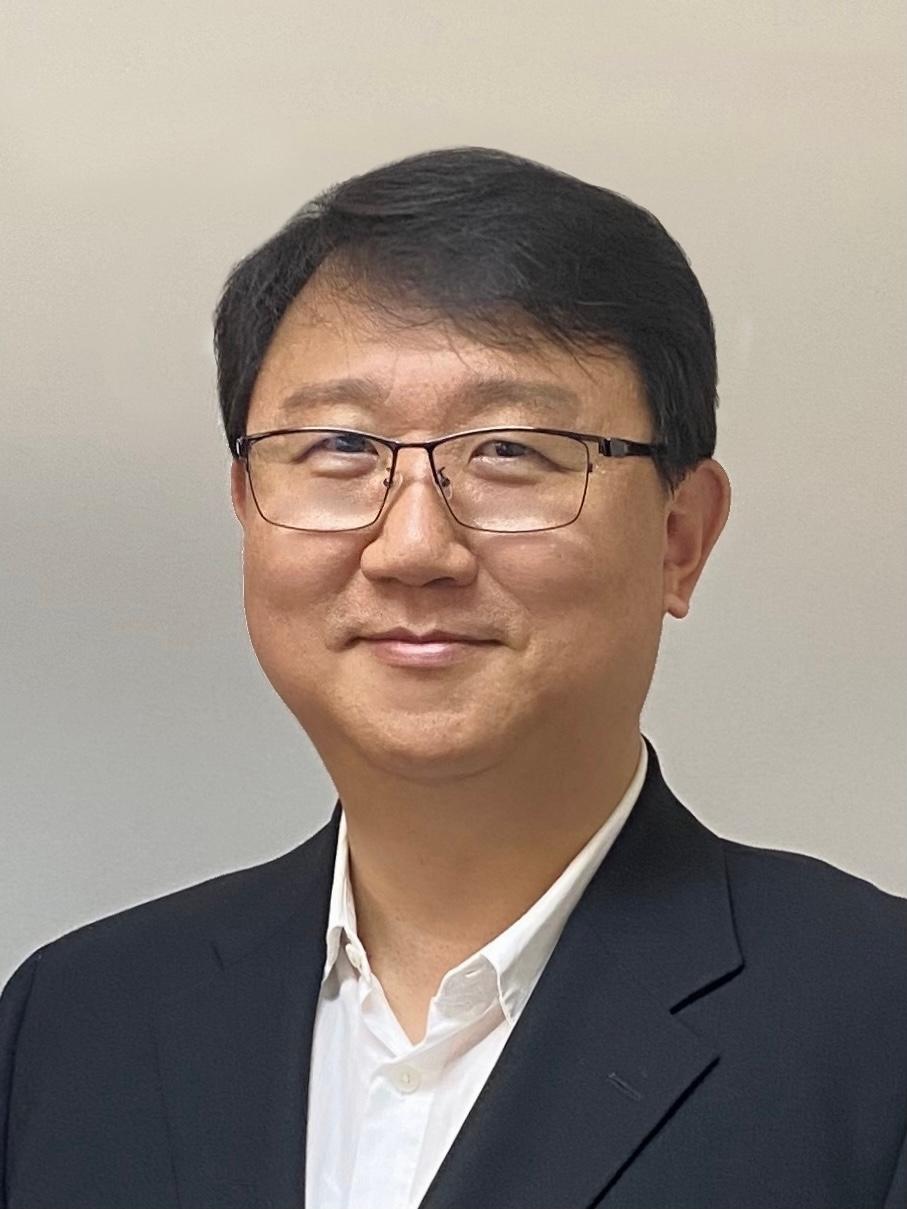}}]{Seung-Hyun Kong} is an Associate Professor in the CCS Graduate School of Mobility, Korea Advanced Institute of Science and Technology (KAIST), where he has been a faculty member since 2010. He received a B.S. degree in Electronics Engineering from Sogang University, Seoul, Korea, in 1992, an M.S. degree in Electrical and Computer Engineering from Polytechnic University (now NYU), New York, in 1994, and a Ph.D. degree in Aeronautics and Astronautics from Stanford University, Palo Alto, in 2005. From 1997 to 2004 and from 2006 to 2010, he was with companies, including Samsung Electronics (Telecommunication Research Center) and Nexpilot, both in Korea, Polaris Wireless in San Jose, and Qualcomm (Corporate R\&D) in San Diego, for advanced R\&D in mobile communication systems, wireless positioning, and assisted GNSS. His current research interest includes deep neural networks for perception and localization using 4D radar and sensor fusion, reinforcement and imitation learning algorithms for end-to-end autonomous driving, and neural network compression for embedded systems. He has authored more than 100 papers in peer-reviewed journals and conference proceedings and 14 patents, and his research group won the President's Award in the 2018 International Student Autonomous Driving Competition from the Korean government. He served as a Program co-Chair of IEEE ITSC 2019, New Zealand, and the General Chair of IEEE IV 2024, Korea, and he has served as an Associate Editor of IEEE T-ITS since 2017.
\end{IEEEbiography}

\setcounter{section}{0}
\setcounter{table}{0}
\setcounter{figure}{0}
\renewcommand{\thesection}{S-\Roman{section}}
\renewcommand{\thesectiondis}{S-\Roman{section}.}
\renewcommand{\thetable}{S-\Roman{table}}
\renewcommand{\thefigure}{S\arabic{figure}}

\twocolumn[{%
\begin{center}
{\LARGE Supplementary Material}\\[0.6em]
{\large RadarGSOcc: Doppler-Aware 3D Semantic Occupancy Prediction Using 4D Radar Tensor with Geometric-Semantic Feature Extraction}
\end{center}
\vspace{1.0em}}]

This supplementary material provides details of the K-Radar 3DSemOcc benchmark (scene composition, class statistics, and label verification), the definition of the Doppler mean/variance channels, implementation details of the deformable attention modules, results for all semantic classes, the environment-wise class-distribution statistics that motivate the EBKL proxy metric, per-run results, and the data behind the mIoU--EBKL correlation. Sections, tables, and figures are prefixed with ``S'' to distinguish them from those of the main paper; references are listed in the main paper.

\section{K-Radar 3DSemOcc Benchmark Details}
\label{supp:benchmark}

\subsection{Scene Composition}
\label{supp:scenes}
The benchmark is built from 12 K-Radar sequences: scenes 2, 5, 8, 9, 10, 14, 16, 17, and 20 for training (5,910 frames) and scenes 3, 6, and 15 for testing (1,781 frames). All labeled scenes are recorded under normal weather conditions. Table~\ref{tab:supp_scenes} lists the environment, time of day, and number of frames of each scene, together with the unlabeled adverse-weather sequences used in Section~\ref{sec:adverse} of the main paper, which are disjoint from the labeled scenes. EBKL is not reported for the fog sequence (scene 38), which is used only in the voxel-ratio analysis (Table~\ref{tab:weather}).

\begin{table}[t]
\centering
\caption{Composition of the K-Radar 3DSemOcc benchmark and the unlabeled adverse-weather sequences.}
\label{tab:supp_scenes}
\setlength{\tabcolsep}{4pt}
\renewcommand{\arraystretch}{1.1}
\begin{tabular}{c l l l l r}
\hline
\textbf{Scene} & \textbf{Split} & \textbf{Weather} & \textbf{Environment} & \textbf{Time} & \textbf{\#Frames} \\
\hline \hline
2  & Train & Normal & Highway & Night & 462 \\
5  & Train & Normal & Urban & Day & 595 \\
8  & Train & Normal & University & Night & 567 \\
9  & Train & Normal & Highway & Day & 833 \\
10 & Train & Normal & Highway & Day & 1,130 \\
14 & Train & Normal & Urban & Day & 595 \\
16 & Train & Normal & University & Day & 578 \\
17 & Train & Normal & University & Day & 593 \\
20 & Train & Normal & Urban & Day & 557 \\
\hline
\multicolumn{5}{l}{Training total} & 5,910 \\
\hline
3  & Test  & Normal & Highway & Night & 596 \\
6  & Test  & Normal & Urban & Night & 594 \\
15 & Test  & Normal & Urban & Day & 591 \\
\hline
\multicolumn{5}{l}{Test total} & 1,781 \\
\hline
23 & Adverse & Rain & Urban & Night & 598 \\
24 & Adverse & Rain & Urban & Night & 598 \\
27 & Adverse & Sleet & Suburban & Day & 597 \\
34 & Adverse & Rain & Suburban & Night & 598 \\
38 & Adverse & Fog & Mountain & Day & 597 \\
46 & Adverse & Heavy snow & Highway & Night & 579 \\
48 & Adverse & Light snow & Highway & Night & 569 \\
50 & Adverse & Sleet & Highway & Night & 590 \\
51 & Adverse & Sleet & Highway & Night & 589 \\
54 & Adverse & Heavy snow & Urban & Day & 573 \\
55 & Adverse & Heavy snow & Urban & Day & 572 \\
\hline
\end{tabular}
\end{table}

\subsection{Class Statistics}
\label{supp:classes}
Tables~\ref{tab:supp_class_before} and \ref{tab:supp_class_after} report the number and percentage of ground-truth voxels per class in the training and test splits before and after the label refinement of Section~\ref{sec:benchmark}; all experiments use the refined labels. Small foreground classes (traffic cone, pedestrian, motorcycle, and bicycle) each account for less than 0.1\% of all voxels, which explains the near-zero IoU of all methods on these classes and motivates the use of W-mIoU in the main paper. Refinement increases the number of voxels of the moving-object classes (car, truck/bus, trailer, pedestrian, motorcycle, and bicycle) in the training split from 364,897 to 767,518.

\begin{table}[t]
\centering
\caption{Ground-truth voxel statistics per class before label refinement (label-confidence threshold 0.8).}
\label{tab:supp_class_before}
\footnotesize
\setlength{\tabcolsep}{4pt}
\renewcommand{\arraystretch}{1.1}
\begin{tabular}{l r r r r}
\hline
\textbf{Class} & \textbf{\#Voxels (train)} & \textbf{\%} & \textbf{\#Voxels (test)} & \textbf{\%} \\
\hline \hline
Drivable area  & 4,183,019 & 24.12 & 1,216,791 & 26.67 \\
Vegetation     & 8,536,616 & 49.22 & 2,068,116 & 45.34 \\
Manmade        & 1,776,829 & 10.24 & 382,270 & 8.38 \\
Terrain        & 2,056,773 & 11.86 & 710,776 & 15.58 \\
Sidewalk       & 406,188 & 2.34 & 91,606 & 2.01 \\
Other flat     & 12,819 & 0.07 & 6,723 & 0.15 \\
Barrier        & 5,678 & 0.03 & 1,629 & 0.04 \\
Car            & 161,384 & 0.93 & 39,778 & 0.87 \\
Truck/bus      & 171,656 & 0.99 & 36,146 & 0.79 \\
Trailer        & 17,235 & 0.10 & 3,111 & 0.07 \\
Traffic cone   & 1,574 & 0.01 & 710 & 0.02 \\
Pedestrian     & 12,218 & 0.07 & 3,818 & 0.08 \\
Motorcycle     & 2,195 & 0.01 & 145 & $<$0.01 \\
Bicycle        & 209 & $<$0.01 & 35 & $<$0.01 \\
\hline
Total          & 17,344,393 & 100 & 4,561,654 & 100 \\
\hline
\end{tabular}
\end{table}

\begin{table}[t]
\centering
\caption{Ground-truth voxel statistics per class after label refinement (label-confidence threshold 0.8), used in all experiments.}
\label{tab:supp_class_after}
\footnotesize
\setlength{\tabcolsep}{4pt}
\renewcommand{\arraystretch}{1.1}
\begin{tabular}{l r r r r}
\hline
\textbf{Class} & \textbf{\#Voxels (train)} & \textbf{\%} & \textbf{\#Voxels (test)} & \textbf{\%} \\
\hline \hline
Drivable area  & 3,738,578 & 22.94 & 1,206,770 & 25.95 \\
Vegetation     & 7,977,664 & 48.96 & 2,048,675 & 44.06 \\
Manmade        & 1,526,764 & 9.37 & 372,645 & 8.01 \\
Terrain        & 1,888,009 & 11.59 & 704,612 & 15.15 \\
Sidewalk       & 378,405 & 2.32 & 91,151 & 1.96 \\
Other flat     & 12,065 & 0.07 & 6,518 & 0.14 \\
Barrier        & 4,828 & 0.03 & 1,477 & 0.03 \\
Car            & 412,504 & 2.53 & 75,112 & 1.62 \\
Truck/bus      & 318,448 & 1.95 & 137,398 & 2.95 \\
Trailer        & 10,495 & 0.06 & 1,203 & 0.03 \\
Traffic cone   & 1,324 & 0.01 & 702 & 0.02 \\
Pedestrian     & 14,216 & 0.09 & 2,867 & 0.06 \\
Motorcycle     & 3,593 & 0.02 & 699 & 0.02 \\
Bicycle        & 8,262 & 0.05 & 35 & $<$0.01 \\
\hline
Total          & 16,295,155 & 100 & 4,649,864 & 100 \\
\hline
\end{tabular}
\end{table}

The benchmark therefore has 15 classes in total: the 14 semantic classes listed in Table~\ref{tab:supp_class_after} and the empty class. All IoU-based metrics in the main paper are computed over the 14 semantic classes.

\subsection{Label Verification}
\label{supp:verification}
\begin{figure*}[t]
\centering
\includegraphics[width=1.8\columnwidth]{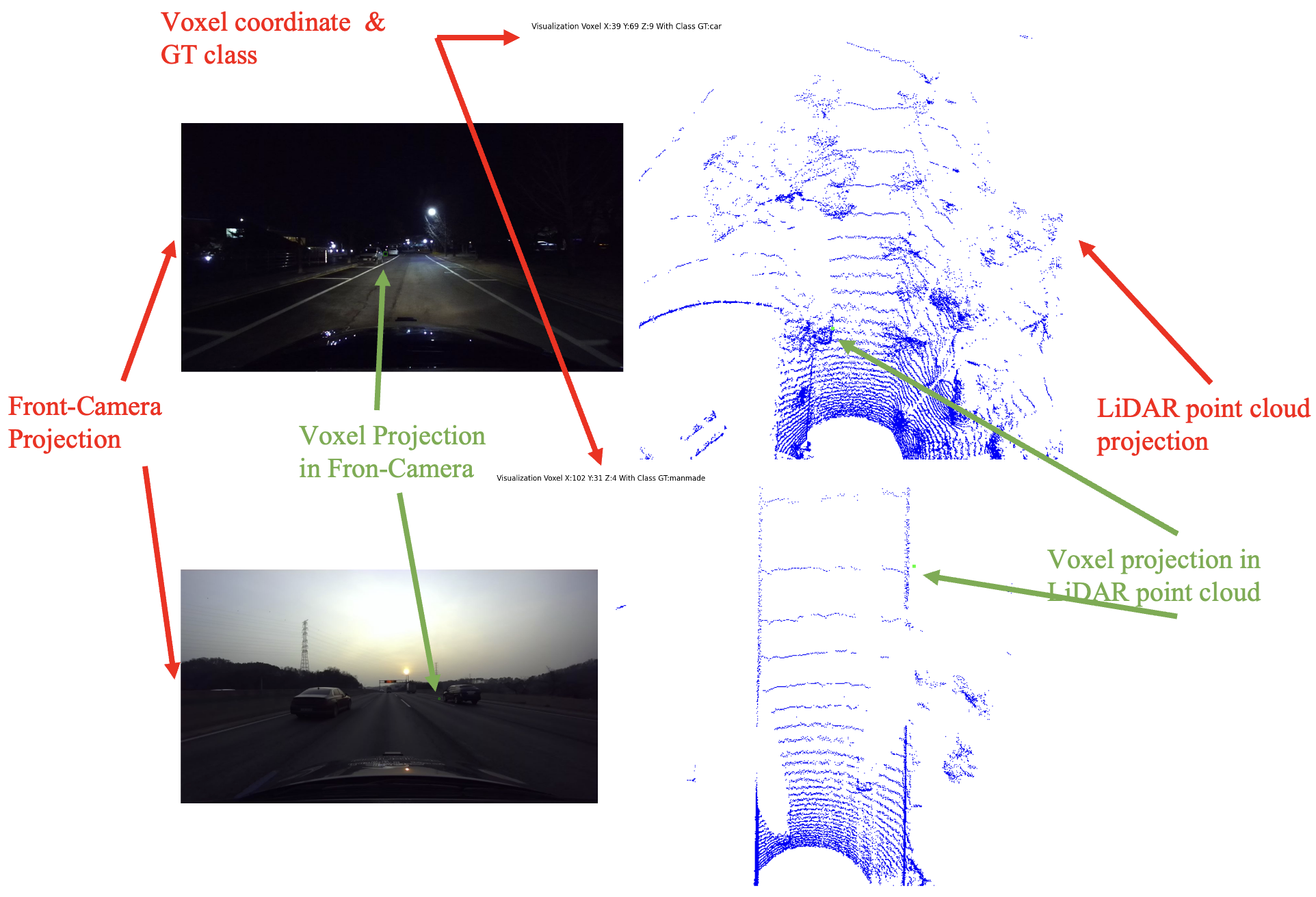}
\caption{Manual verification of the K-Radar 3DSemOcc benchmark. Sampled voxels (green 3D boxes) are projected onto the front-camera image (left) and the LiDAR bird's-eye view (right): (top) car voxels in a night scene and (bottom) manmade voxels in a day scene.}
\label{fig:supp_voxel_check}
\end{figure*}

To assess the quality of the refined pseudo-labels, 525 non-empty voxels were randomly sampled from the benchmark (label-confidence threshold 0.8) and verified manually. Each sampled voxel is drawn as a green 3D box and projected onto the front-camera image and onto the bird's-eye view of the LiDAR point cloud of the same frame (Fig.~\ref{fig:supp_voxel_check}); a voxel is counted as correct when its semantic label agrees with the object or surface visible at that location in both views. The overall accuracy is 91.8\% (482 of 525), and per-class results are given in Table~\ref{tab:supp_verification}; terrain has the lowest accuracy (69.7\%).

\begin{table}[t]
\centering
\caption{Manual verification of 525 randomly sampled non-empty voxels.}
\label{tab:supp_verification}
\setlength{\tabcolsep}{5pt}
\renewcommand{\arraystretch}{1.1}
\begin{tabular}{l r r r}
\hline
\textbf{Class} & \textbf{\#Sampled} & \textbf{\#Correct} & \textbf{Accuracy (\%)} \\
\hline \hline
Drivable area  & 126 & 125 & 99.2 \\
Vegetation     & 191 & 188 & 98.4 \\
Manmade        & 66 & 64 & 97.0 \\
Terrain        & 99 & 69 & 69.7 \\
Car            & 16 & 14 & 87.5 \\
Truck/bus      & 5 & 4 & 80.0 \\
Other classes  & 22 & 18 & 81.8 \\
\hline
All            & 525 & 482 & 91.8 \\
\hline
\end{tabular}
\end{table}

\section{Doppler Mean and Variance Channels}
\label{supp:doppler}
In addition to the top-3 power values and their Doppler indices, the 8-channel sparse tensor contains the mean and the variance of the power along the Doppler dimension of each spatial element, following RadarOcc~\cite{ding2024radarocc}. For a spatial element with Doppler bins $d = 1,\dots,N_D$ and corresponding power $p_d$, they are defined as
\begin{align}
\mu_D &= \frac{1}{N_D}\sum_{d=1}^{N_D} p_d, \\
\sigma_D^2 &= \frac{1}{N_D}\sum_{d=1}^{N_D} \left(p_{d}-\mu_D\right)^2.
\end{align}

\section{Implementation Details of the Deformable Attention Modules}
\label{supp:deformable}
RadarGSOcc uses 3D deformable attention in three places: the deformable self-attention in the geometric and semantic branches, the geometric-semantic cross-attention fusion, and the geometric feature refinement in the occupancy head. Table~\ref{tab:supp_deform} summarizes their hyperparameters.

\begin{table}[t]
\centering
\caption{Hyperparameters of the deformable attention modules.}
\label{tab:supp_deform}
\setlength{\tabcolsep}{4pt}
\renewcommand{\arraystretch}{1.1}
\begin{tabular}{l c c c}
\hline
 & \textbf{Self-attn.} & \textbf{Fusion} & \textbf{Refinement} \\
\hline \hline
Feature dimension $dim$      & 96 & 96 & 192 \\
Number of heads              & 8 & 8 & 8 \\
Sampling points per query    & 8 & 8 & 8 \\
Number of layers             & 1 & 1 & 1 \\
Sampling coordinates         & $[0,1]$ & $[0,1]$ & $[0,1]$ \\
\hline
\end{tabular}
\end{table}

The geometric and semantic branches use a feature dimension of 96 each, so that their concatenation matches the 192-dimensional occupancy features used in the refinement module and in RadarOcc. Following RadarOcc, the reference points are initialized uniformly along the range, azimuth, and elevation dimensions of the spherical grid, and the sampling locations $ref + \Delta$ are expressed in normalized coordinates in $[0, 1]$ along each axis (0 and 1 denote the two ends of the axis). Features at the sampled locations are obtained by trilinear interpolation, and the softmax in the attention equations of Section~\ref{sec:gsfe} and Section~\ref{sec:occhead} of the main paper is taken over the sampled locations.

\section{Results for All Semantic Classes}
\label{supp:full_results}
Table~\ref{tab:supp_full} extends Table~\ref{tab:main} of the main paper to all semantic classes. Except for OpenOcc-C, the camera-based, radar point cloud-based, and camera--radar fusion baselines obtain zero IoU on all seven classes on the right.

\begin{table*}[t]
\centering
\caption{3D semantic occupancy results on the K-Radar 3DSemOcc benchmark for all classes (IoU in \%, test split).}
\label{tab:supp_full}
\scriptsize
\setlength{\tabcolsep}{2pt}
\renewcommand{\arraystretch}{1.1}
\begin{tabular}{l c | ccc | cccccccccccccc}
\hline
\textbf{Method} & \textbf{Sensor}
& \textbf{IoU} & \textbf{mIoU} & \textbf{W-mIoU}
& \textbf{DA} & \textbf{Veg} & \textbf{Man} & \textbf{Car} & \textbf{Truck}
& \textbf{Ter} & \textbf{Side} & \textbf{Other} & \textbf{Trail} & \textbf{Cone}
& \textbf{Ped} & \textbf{Moto} & \textbf{Bike} & \textbf{Bar} \\
\hline \hline
CONet-C~\cite{wang2023openoccupancylargescalebenchmark} & Cam
& 16.9 & 3.2 & 11.3
& 27.4 & 4.2 & 0.0 & 0.6 & 0.0 & 11.4 & 0.0 & 0.0 & 0.0 & 0.0 & 0.0 & 0.0 & 0.0 & 0.0 \\
BEVFormer~\cite{li2024bevformer} & Cam
& 27.3 & 4.1 & 16.0
& 34.0 & 10.4 & 0.1 & 0.0 & 0.0 & 11.9 & 0.5 & 0.0 & 0.0 & 0.0 & 0.0 & 0.0 & 0.0 & 0.0 \\
OpenOcc-C~\cite{wang2023openoccupancylargescalebenchmark} & Cam
& 30.0 & 5.4 & 17.1
& 34.7 & 17.3 & 13.7 & 5.9 & 1.8 & 19.0 & 0.9 & 1.6 & 0.0 & 0.0 & 0.0 & 0.0 & 0.0 & 1.5 \\
OpenOcc-R~\cite{wang2023openoccupancylargescalebenchmark} & Radar
& 26.0 & 4.0 & 14.7
& 34.0 & 8.8 & 5.1 & 1.2 & 0.0 & 8.9 & 0.8 & 0.0 & 0.0 & 0.0 & 0.0 & 0.0 & 0.0 & 0.0 \\
OpenOcc-RC~\cite{wang2023openoccupancylargescalebenchmark} & R+Cam
& 25.2 & 3.8 & 14.0
& 34.9 & 5.6 & 0.8 & 0.0 & 0.1 & 11.0 & 1.4 & 0.0 & 0.0 & 0.0 & 0.0 & 0.0 & 0.0 & 0.0 \\
4DRC-Occ~\cite{ninfa20264drc} & R+Cam
& 30.2 & 4.6 & 17.6
& 35.2 & 12.8 & 0.1 & 0.2 & 0.0 & 15.3 & 1.9 & 0.0 & 0.0 & 0.0 & 0.0 & 0.0 & 0.0 & 0.0 \\
OpenOcc-L (64)~\cite{wang2023openoccupancylargescalebenchmark} & LiDAR
& 53.1 & 10.9 & 36.1
& 50.5 & 39.9 & 15.4 & 3.3 & 2.4 & 26.1 & 9.1 & 6.4 & 0.0 & 0.0 & 0.2 & 0.0 & 0.0 & 0.0 \\
OpenOcc-L (128)~\cite{wang2023openoccupancylargescalebenchmark} & LiDAR
& 54.3 & 11.1 & 38.9
& 51.5 & 43.0 & 15.4 & 1.5 & 3.3 & 30.3 & 6.4 & 3.8 & 0.2 & 0.0 & 0.2 & 0.0 & 0.0 & 0.8 \\
RadarOcc~\cite{ding2024radarocc} & RadarT
& 36.4 & 6.7 & 21.9
& 40.4 & 16.3 & 10.3 & 6.3 & 0.6 & 18.0 & 2.4 & 0.7 & 0.0 & 0.0 & 0.1 & 0.0 & 0.0 & 0.5 \\
\hline
\textbf{RadarGSOcc (Ours)} & RadarT
& 38.2 & 7.6 & 25.5
& 42.2 & 21.2 & 11.6 & 0.6 & 4.7 & 23.9 & 2.5 & 0.2 & 0.0 & 0.0 & 0.0 & 0.0 & 0.0 & 0.6 \\
+ w/o overlap & RadarT
& 35.7 & 6.8 & 21.8
& 34.5 & 17.5 & 11.9 & 0.6 & 2.1 & 21.2 & 4.9 & 0.1 & 1.3 & 0.0 & 0.8 & 0.2 & 0.0 & 0.4 \\
\hline
\end{tabular}
\end{table*}

\section{Environment-Wise Class Distribution for EBKL}
\label{supp:ebkl}
\begin{figure}[t]
\centering
\includegraphics[width=0.85\columnwidth]{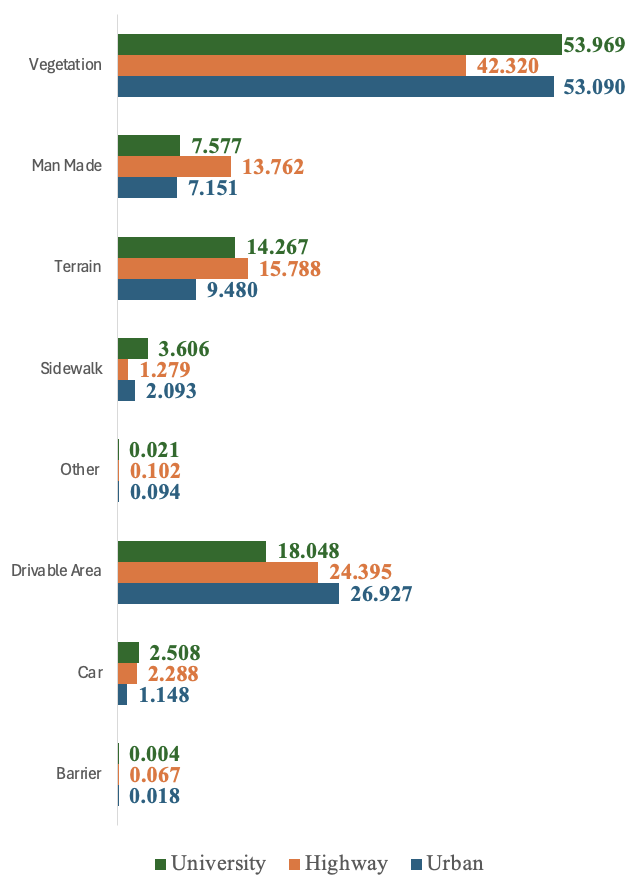}
\caption{Normalized ground-truth class distributions (\%) of the normal-weather scenes, grouped by environment, for the eight classes shown.}
\label{fig:supp_ebkl_dist}
\end{figure}

EBKL assumes that, within the same environment type (urban, university campus, or highway), the class distribution of the scene is relatively stable across scenes and weather conditions. Fig.~\ref{fig:supp_ebkl_dist} shows the normalized ground-truth class distribution of the normal-weather scenes grouped by environment, and Table~\ref{tab:supp_ebkl_within} reports the mean EBKL between the individual scenes of each environment (rows) and each environment-level reference distribution (columns). Highway scenes are clearly closer to the highway reference (0.015) than to the urban and university references (0.032 and 0.036), and university scenes are closest to their own reference (0.008). Urban scenes, however, are slightly closer to the university reference (0.014) than to their own (0.017), so the urban and university-campus distributions are not well separated; in K-Radar, EBKL therefore mainly distinguishes highway from non-highway scenes.

\begin{table}[t]
\centering
\caption{Mean EBKL between scene-level and environment-level class distributions (ground truth, normal weather).}
\label{tab:supp_ebkl_within}
\setlength{\tabcolsep}{5pt}
\renewcommand{\arraystretch}{1.1}
\begin{tabular}{l c c c}
\hline
\textbf{Scenes $\backslash$ Reference} & \textbf{Urban} & \textbf{University} & \textbf{Highway} \\
\hline \hline
Urban scenes      & 0.017 & 0.014 & 0.030 \\
University scenes & 0.015 & 0.008 & 0.035 \\
Highway scenes    & 0.032 & 0.036 & 0.015 \\
\hline
\end{tabular}
\end{table}

\section{Per-Run Results}
\label{supp:seeds}
Table~\ref{tab:supp_seeds} lists the results of the three training runs of RadarOcc and RadarGSOcc that are summarized in Section~\ref{sec:seeds} of the main paper. The first run of each model was trained without a fixed random seed and is the run reported in Table~\ref{tab:main} and the other tables of the main paper; the second and third runs use seeds 1111 and 2222. The means and standard deviations are computed from the unrounded values.

\begin{table}[t]
\centering
\caption{Results of the three training runs of RadarOcc and RadarGSOcc on the K-Radar 3DSemOcc benchmark (test split).}
\label{tab:supp_seeds}
\setlength{\tabcolsep}{4pt}
\renewcommand{\arraystretch}{1.1}
\begin{tabular}{l l | c c c}
\hline
\textbf{Model} & \textbf{Run (seed)} & \textbf{IoU} & \textbf{mIoU} & \textbf{W-mIoU} \\
\hline \hline
\multirow{4}{*}{RadarOcc}   & 1 (not fixed) & 36.4 & 6.7 & 21.9 \\
                            & 2 (1111) & 36.3 & 6.4 & 20.1 \\
                            & 3 (2222) & 36.0 & 7.0 & 20.8 \\
                            & Mean $\pm$ std & 36.26$\pm$0.23 & 6.72$\pm$0.28 & 20.93$\pm$0.90 \\
\hline
\multirow{4}{*}{RadarGSOcc} & 1 (not fixed) & 38.2 & 7.6 & 25.5 \\
                            & 2 (1111) & 39.3 & 7.0 & 24.6 \\
                            & 3 (2222) & 34.8 & 6.8 & 21.4 \\
                            & Mean $\pm$ std & 37.43$\pm$2.34 & 7.14$\pm$0.41 & 23.83$\pm$2.15 \\
\hline
\end{tabular}
\end{table}

\section{Data Behind the mIoU--EBKL Correlation}
\label{supp:corr}
Table~\ref{tab:supp_corr} lists the mIoU (Table~\ref{tab:main}) and the EBKL under normal weather of the nine models used for Fig.~\ref{fig:miou_ebkl} of the main paper. Over these nine models, the Pearson correlation coefficient between mIoU and EBKL is $-0.87$ (Spearman rank correlation $-0.95$): models with higher mIoU have lower EBKL.

\begin{table}[t]
\centering
\caption{mIoU and EBKL (normal weather, test split) of the nine models in Fig.~\ref{fig:miou_ebkl}.}
\label{tab:supp_corr}
\setlength{\tabcolsep}{5pt}
\renewcommand{\arraystretch}{1.1}
\begin{tabular}{l c c}
\hline
\textbf{Model} & \textbf{mIoU (\%)} & \textbf{EBKL} \\
\hline \hline
CONet-C~\cite{wang2023openoccupancylargescalebenchmark} & 3.2 & 0.093 \\
BEVFormer~\cite{li2024bevformer} & 4.1 & 0.072 \\
OpenOcc-C~\cite{wang2023openoccupancylargescalebenchmark} & 5.4 & 0.0536 \\
OpenOcc-R~\cite{wang2023openoccupancylargescalebenchmark} & 4.0 & 0.054 \\
OpenOcc-RC~\cite{wang2023openoccupancylargescalebenchmark} & 3.8 & 0.075 \\
4DRC-Occ~\cite{ninfa20264drc} & 4.6 & 0.0543 \\
OpenOcc-L (64)~\cite{wang2023openoccupancylargescalebenchmark} & 10.9 & 0.0214 \\
RadarOcc~\cite{ding2024radarocc} & 6.7 & 0.0322 \\
RadarGSOcc (Ours) & 7.6 & 0.0218 \\
\hline
\end{tabular}
\end{table}

\end{document}